%% file: main.tex
\documentclass[journal]{IEEEtran}

\ifCLASSINFOpdf
\else
\fi

\usepackage[colorlinks,linkcolor=blue]{hyperref}
\usepackage{graphicx}
\usepackage[numbers]{natbib}

\begin{document}

\title{Making Images Real Again: A Comprehensive Survey on Deep Image Composition}

\author{Li Niu*, Wenyan Cong, Liu Liu, Yan Hong, Bo Zhang, Jing Liang, Liqing Zhang
\thanks{Li Niu, Wenyan Cong, Liu Liu, Yan Hong, Bo Zhang, Jing Liang, and Liqing Zhang are with MOE Key Lab of Artificial Intelligence, Department of Computer Science and Engineering Shanghai Jiao Tong University, Shanghai, China (email: \{ustcnewl,plcwyam17320,shirlley,Hy2628982280, bo-zhang,leungjing, lqzhang\}@sjtu.edu.cn).}
\thanks{* means the corresponding author.}
}

\maketitle

\input{sections/abstract}

\IEEEpeerreviewmaketitle

\input{sections/introduction}

\input{sections/object_placement}

\input{sections/image_blending}

\input{sections/image_harmonization}
\input{sections/shadow_generation}

\input{sections/reflection_generation}
\input{sections/generative_composition}

\input{sections/foreground_object_search}

\input{sections/conclusion}

\ifCLASSOPTIONcaptionsoff
  \newpage
\fi

\bibliographystyle{plainnat}
\bibliography{composition}

\end{document}

%% file: sections/abstract.tex
\begin{abstract}
As a common image editing operation, image composition/compositing, which is also called object/subject insertion/addition/compositing, aims to combine the foreground from one image and another background image to produce a composite image. However, there are many issues that could make the composite images unrealistic. These issues can be summarized as the inconsistency between foreground and background, which includes appearance inconsistency (\emph{e.g.}, incompatible illumination), geometry inconsistency (\emph{e.g.}, unreasonable size), and semantic inconsistency (\emph{e.g.}, mismatched semantic context). The image composition task could be decomposed into multiple sub-tasks, in which each sub-task targets one or more issues. Specifically, object placement aims to find reasonable scale, location, and shape for the foreground. Image blending aims to address the unnatural boundary between foreground and background. Image harmonization aims to adjust the illumination statistics of foreground. Shadow (\emph{resp.}, reflection) generation aims to generate plausible shadow (\emph{resp.}, reflection) for the foreground. These sub-tasks can be executed sequentially or in parallel to acquire realistic composite images. To the best of our knowledge, there is no previous survey on image composition. In this paper, we conduct a comprehensive survey over the sub-tasks and combined task of image composition. For each one, we summarize the existing methods, available datasets, and common evaluation metrics. Datasets and codes for image composition are summarized at \href{https://github.com/bcmi/Awesome-Object-Insertion}{https://github.com/bcmi/Awesome-Image-Composition}. We have also contributed the first image composition toolbox: libcom \href{https://github.com/bcmi/libcom}{https://github.com/bcmi/libcom}, which assembles 10+ image-composition-related functions (\emph{e.g.}, image blending, image harmonization, object placement, shadow/reflection generation, generative composition). The ultimate goal of this toolbox is to solve all image composition problems with simple \emph{`import libcom'}. Based on libcom toolbox, we also develop an online image composition workbench \href{https://libcom.ustcnewly.com}{https://libcom.ustcnewly.com}. 
\end{abstract}

%% file: sections/introduction.tex
\section{Introduction}\label{sec:intro}

Image composition/compositing~\cite{STGANChen2018,chen2019toward,zhan2020adversarial,TFICON}, which is also called object/subject insertion/addition/compositing in some literature~\cite{singh2023smartmask,winter2024objectdrop,bhattad2020cutandpaste,objectstitch}, aims to combine the foreground from one image and another background image to form a composite image. More generally, image composition can be used for combining multiple visual elements from different sources to construct a new image, which is a common image editing operation. 
After compositing a new image with foreground and background,  there exist many issues that could make the composite image unrealistic and thus significantly degrade its quality. These issues can be summarized as the inconsistency between foreground and background, which can be divided into appearance inconsistency, geometric inconsistency, and semantic inconsistency. Each type of inconsistency involves a number of issues to be solved. Image composition task could be decomposed into multiple sub-tasks, in which each sub-task 
targets at one or more issues. Next, we will introduce each type of inconsistency one by one.

\begin{figure}[t]
\begin{center}
\includegraphics[width=.95\linewidth]{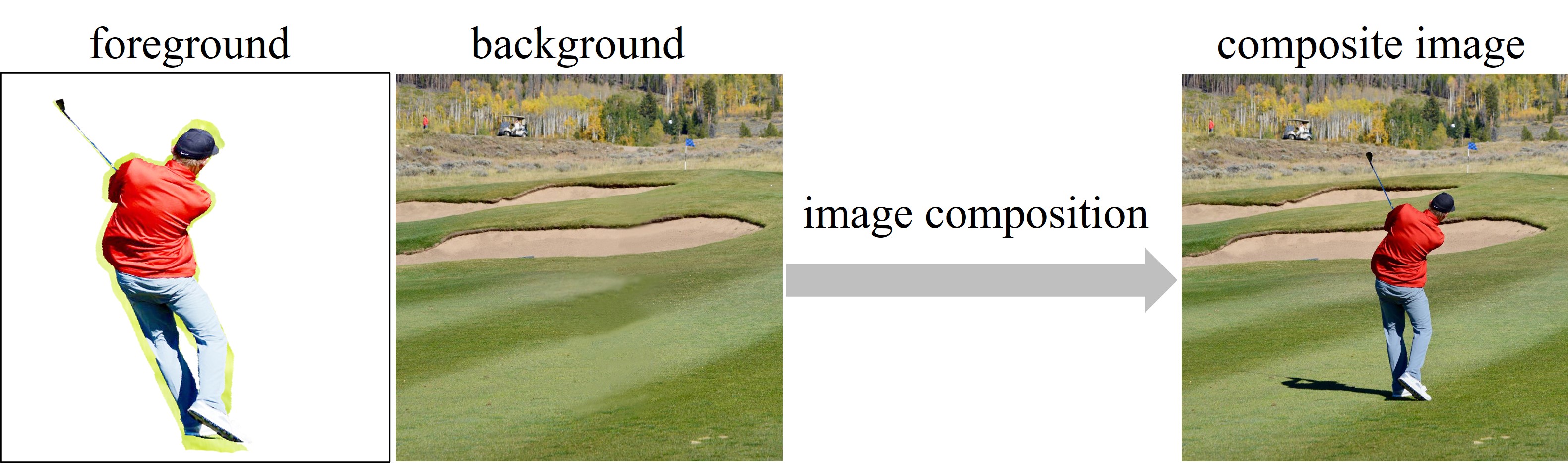}
\end{center}
\caption{Image composition aims to combine the foreground object and the background image to generate a realistic composite image. 
}
\label{fig:overview}
\end{figure}

\begin{figure*}[t]
\begin{center}
\includegraphics[width=.95\linewidth]{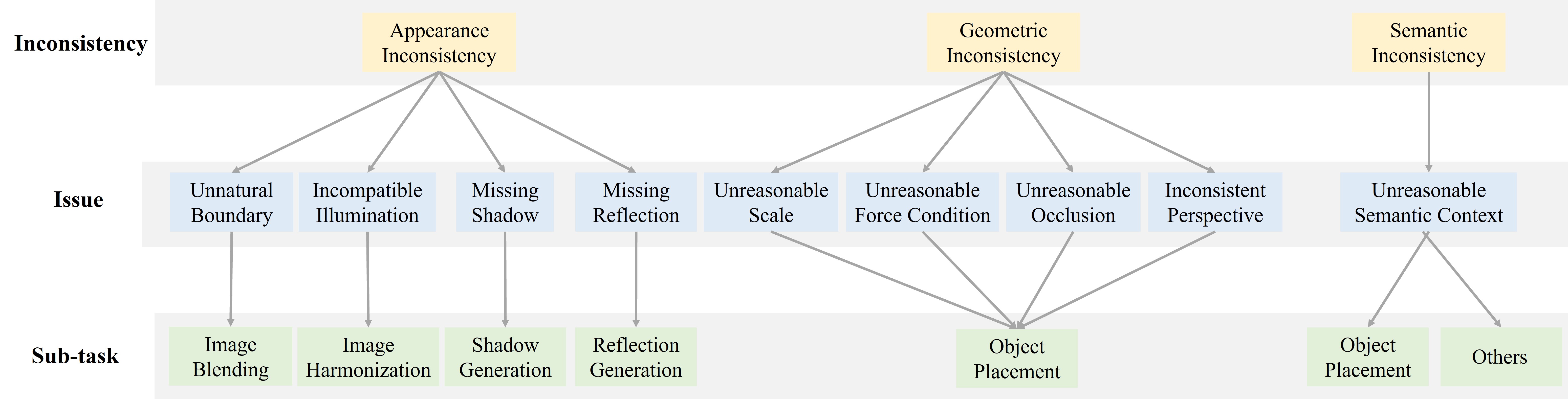}
\end{center}
\caption{The quality of composite image is degraded by the appearance inconsistency, geometric inconsistency, and semantic inconsistency. Each type of inconsistency involves a number of issues. Each sub-task targets one or more issues. 
}
\label{fig:summary}
\end{figure*}

\begin{figure*}[t]
\begin{center}
\includegraphics[width=.95\linewidth]{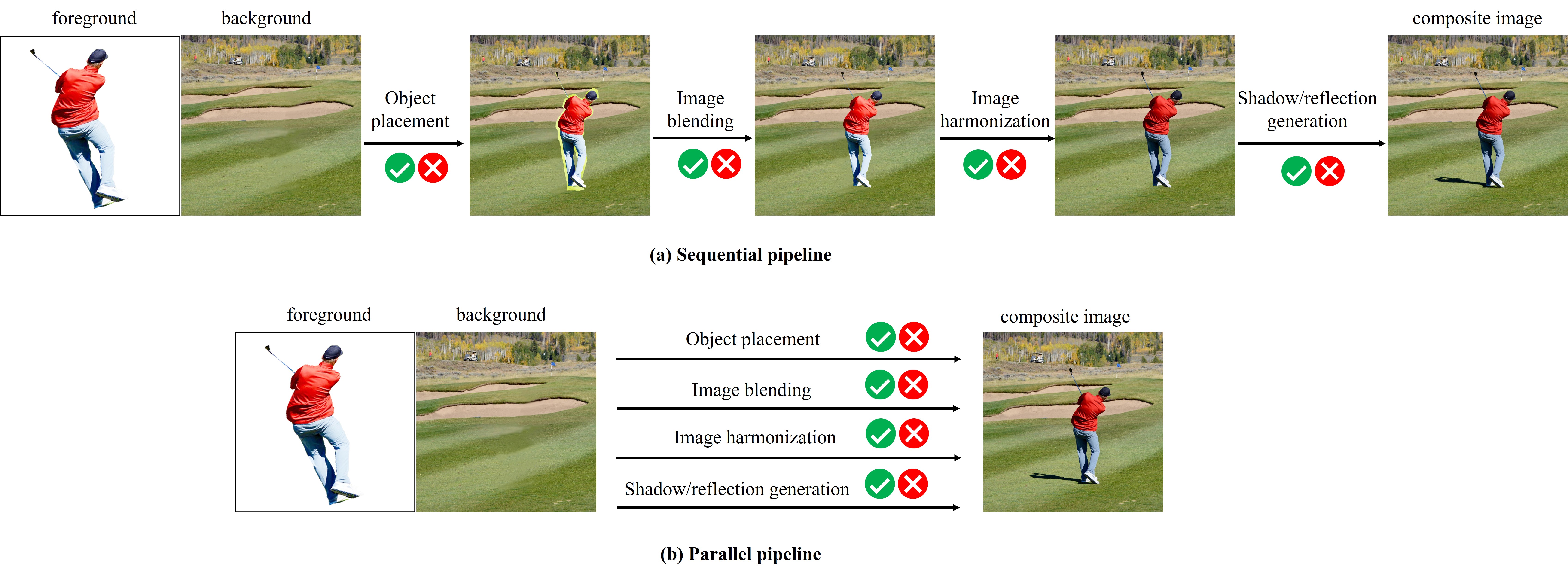}
\end{center}
\caption{Previous works perform multiple sub-tasks (\emph{e.g.}, object placement, image blending, image harmonization, shadow/reflection generation) sequentially or in parallel to achieve the goal of image composition. 
}
\label{fig:flowchart}
\end{figure*}

\begin{table*} [t]
\caption{The issues to be solved in image composition task and the corresponding deep learning methods to solve these issue. Note that some methods only focus on one issue while some methods attempt to solve multiple issues simultaneously. ``Boundary" means refining the boundary between foreground and background. ``Appearance" means adjusting the illumination of foreground. ``Shadow" means generating shadow for the foreground. ``Reflection" means generating reflection for the foreground.  ``Geometry" means seeking  reasonable location, size, and shape for the foreground considering geometric constraints. ``Occlusion" means coping with the unreasonable occlusion. ``Semantics" means finding suitable semantic context for the foreground. }
\label{tab:method_summary}
\centering
\normalsize
\setlength{\tabcolsep}{5pt}
\begin{tabular}{|ccccccc|p{6cm}|}
\hline
boundary & appearance &  shadow & reflection & geometry & occlusion & semantics & methods\\ \hline
+ &  &  & & & & &  \cite{zhang2021deep} \\ \hline
  & + &  & & & & & \cite{tsai2017deep,DoveNet2020,bargain,sofiiuk2021foreground,intriharm,regionaware,jiang2021ssh,guo2021image,cong2022high,DCCF,Harmonizer,PCTNet,DucoNet,HDNet,ShenICCV2023,GiftNet} \\ \hline
+ & + &  & &  & & & \cite{wu2019gp,zhang2020deep,Xing0WZC22,wang2024refined}  \\ \hline
  & + &  &  & + & & + &  \cite{zhan2019spatial}   \\ \hline
  & + & + & &  & & &   \cite{zhan2020adversarial,bao2022deep,zhou2024foreground,yu2024cfdiffusion}  \\ \hline
  &  & + & & & & &   \cite{liu2020arshadowgan,zhang2019shadowgan,hong2021shadow,RdSOBA,DESOBAv2,zhao2025shadow} \\ \hline
    &  &  & + & & & &   \cite{rgdiffusion} \\ \hline
  &  &  &  & + & & + & \cite{STGANChen2018,SyntheticTripathi2019,LearningObjPlaZhang2020,niu2022fast,graco2022,topnet2023,qin2025think,zhou2025bootplace}    \\ \hline
  &  &  &   & + & + & +& \cite{CompositionalGANAzadi2020,zhanhierarchy}\\\hline
  &  &  &  & & + & & \cite{tan2019image,depth_registration,li2026place}\\\hline
+ & + &  &  & + &  & + & \cite{chen2019toward}\\ \hline
+ & + & & & & + &  & \cite{lee2024compose} \\ \hline
+ & + &  &  & +  &  & + & \cite{he2024affordance} \\ \hline
+ & + & + & + & + &  & + & \cite{PBE,objectstitch,zhang2023controlcom,carecom,song2024imprint,anydoor,winter2024objectmate,song2025insert} \\ \hline
+ & + & + & + & + & + & + & \cite{li2024bifr} \\ \hline

\end{tabular}
\end{table*}

The appearance inconsistency includes, but is not limited to: 1) unnatural boundary between foreground and background; 2) incompatible illumination statistics between foreground and background; 3) missing or implausible shadow and reflection of foreground; 4) resolution, sharpness, and noise discrepancy between foreground and background~\cite{li2022bridging}. For the first issue, the foreground is usually extracted using image segmentation \cite{minaee2021image} or matting \cite{xu2017deep,fu2022survey} algorithms. However, the foregrounds may not be precisely delineated, especially at the boundaries. When pasting the foreground with jagged boundaries on the background, there would be obvious color artifacts along the boundary.
To solve this issue, image blending \cite{wu2019gp,zhang2020deep} aims to address the unnatural boundary between foreground and background, so that the foreground could be seamlessly blended with the background. For the second issue, since the foreground and background may be captured in different conditions (\emph{e.g.}, weather, season, time of the day, camera setting), the obtained composite image could lack visual harmony (\emph{e.g.}, foreground captured in the daytime and background captured at night).
To solve this issue, image harmonization \cite{tsai2017deep,xiaodong2019improving,DoveNet2020} aims to adjust the illumination statistics of foreground to make it more compatible with the background, so that the resulting composite image looks more harmonious. For the third issue, when pasting the foreground on the background, the foreground may also affect the background with shadow or reflection. To solve this issue, shadow generation~\cite{liu2020arshadowgan,zhang2019shadowgan,sheng2021ssn} or reflection generation \cite{ma2021neural,winter2024objectdrop,tarres2024thinking} focus on generating plausible shadow or reflection for the foreground according to both foreground and background information. For the fourth issue, the foreground and background may be from two images with different resolutions, blur degrees, and noise patterns. 
The resolution (\emph{resp.}, sharpness, noise) discrepancy between them could be mitigated by using super-resolution~\cite{Wang0H21}, deblurring~\cite{ZhangRLLSYL22}, denoising~\cite{TianFZ0ZL20} techniques.  

The geometric inconsistency includes, but is not limited to: 1) the foreground object is too large or too small; 2) the foreground object does not have reasonable supporting force (\emph{e.g.}, hanging in the air); 3) unreasonable occlusion; 4) inconsistent perspectives between foreground and background. In summary, the location, size, and shape of the foreground may be irrational considering the geometric constraints. 
Object placement  \cite{CompositionalGANAzadi2020,MVCIKAODDDvornik2018,kikuchi2019regularized,SyntheticTripathi2019,LearningObjPlaZhang2020} tends to seek  reasonable location, size, and shape by predicting the foreground transformation to avoid the abovementioned inconsistencies. Previous object placement methods \cite{LearningObjPlaZhang2020,SyntheticTripathi2019} mainly predict simple form of spatial transformation, that is,  shifting and scaling the foreground to achieve reasonable location and size. Some other methods \cite{kikuchi2019regularized,STGANChen2018} predict more general form of spatial transformation (\emph{e.g.}, affine transformation, perspective transformation, thin plate spline transformation) to warp the foreground. In terms of more advanced geometric transformation like view synthesis and pose transfer, we should resort to generative approaches \cite{PBE,objectstitch} to change the viewpoint/pose of the foreground.
When placing the object on the background, unreasonable occlusion may occur. Most previous methods seek reasonable placement to avoid unreasonable occlusions, while some methods \cite{CompositionalGANAzadi2020,zhanhierarchy,tan2019image,li2026place} aim to fix unreasonable occlusion by removing the occluded regions of foreground based on the estimated depth information.  

The semantic inconsistency includes, but is not limited to: 1) the foreground appears at a semantically unreasonable place (\emph{e.g.}, a zebra is placed in the living room); 2) the foreground has unreasonable interactions with other objects or people (\emph{e.g.}, a person is riding a motorbike, but the person and the motorbike are facing towards opposite directions); 3) the background may have semantic impact on the foreground appearance. The semantic inconsistency is judged based on commonsense knowledge, so the cases of semantic inconsistency may be arguable according to subjective judgement. For example, when a car is placed in the water, it can be argued that a car is sinking into the water after a car accident. However, such event has rather low probability compared with commonly seen cases, so we can claim that the car appears at an unreasonable place, which belongs to semantic inconsistency. 
Partial solution to semantic inconsistency falls into the scope of object placement. To be exact, by predicting suitable spatial transformation for the foreground, we can relocate the foreground to a reasonable place or adjust the pose of foreground to make its interactions with environment more convincing. Additionally, the appearance of foreground object may be affected by the background semantically~\cite{zhou2024scene}, which is different from low-level appearance inconsistency (illumination, shadow). For example, a car placed on the snowy ground may be covered by snow. Another example is that a student inserted into a group of students wearing school uniforms should wear the same school uniform. Such semantic appearance editing (included in ``Others" in Fig.~\ref{fig:summary}) is very flexible and challenging, which will not be fully discussed in this survey. 

So far, we have introduced several sub-tasks (\emph{e.g.}, image blending, image harmonization, shadow generation, object placement), in which one sub-task targets one or multiple issues. 
Previous works usually focus on one sub-task or perform multiple sub-tasks sequentially (\emph{i.e.}, image blending followed by image harmonization) as shown in Fig.~\ref{fig:flowchart}(a). The reasonable sequential order is as follows. Given a pair of foreground and background, we first use 
object placement to find suitable scale and location for the foreground, and use image blending  to refine the boundary between foreground and background. Then, we use image harmonization to adjust the foreground illumination and shadow generation to generate plausible shadow for the foreground. 
Recently, as the diffusion models have demonstrated unprecedented generation ability, some works~\cite{PBE,objectstitch} utilize diffusion models to perform multiple sub-tasks (\emph{e.g.}, image blending, image harmonization, view synthesis) parallelly as shown in Fig.~\ref{fig:flowchart}(b). Given a pair of foreground and background with bounding box, they propose one unified model to directly produce the composite image, in which the foreground is  blended seamlessly and harmoniously into the background. 
These methods re-generate the foreground object instead of making restrained adjustments for the foreground object, so we refer to them as generative image composition methods. 
We summarize all the potential issues and the corresponding methods to solve them in Table~\ref{tab:method_summary}. 

Instead of creating realistic composite images from  arbitrary pairs of foregrounds and backgrounds, another solution is seeking for suitable foregrounds from a foreground library, which are compatible with the background in terms of illumination, geometry, and semantics. Finding compatible foregrounds can greatly alleviate the burden of creating realistic composite images, which is complementary with the aforementioned image composition techniques.
This task is called foreground object search \cite{Zhu2022GALATG,Wu2021FinegrainedFR,Li2020InterpretableFO}, which is especially useful when we have a high-quality foreground library with wide coverage. 

Image composition has a broad spectrum of applications in the realm of entertainment, virtual reality, artistic creation, e-commerce \cite{chen2019toward,weng2020misc,WhatWhereZhang2020} and data augmentation \cite{dwibedi2017cut,LSCPRemez2018,ouyang2018pedestrian} for downstream tasks. For example, people can replace the backgrounds of self-portraits and make the obtained images more realistic using image composition techniques~\cite{ValanarasuICLR2023,Xing0WZC22}. Similar application scenarios include virtual conference room or virtual card room. Another example is artistic creation, in which image composition can be used to create fantastic artworks that originally only exist in the imagination. Moreover, image composition could also be used for automatic advertising, which helps advertisers with the product insertion in the background scene~\cite{WhatWhereZhang2020}. When the product is clothes or furniture, this application scenario is also known as virtual try-on or virtual home decoration~\cite{STGANChen2018}. Similarly, advertisement logo compositing \cite{li2019advertisement} targets at embedding some specified logos in target images. The obtained composite images can be taken as design renderings or blueprint to help the designer and the client choose their preferable versions. Additionally, image composition could create synthetic composite images with close data distribution to real images, to augment the training data for downstream tasks like object detection and instance segmentation~\cite{dwibedi2017cut,LSCPRemez2018,ouyang2018pedestrian}. 

In the remainder of this paper, we will elaborate on each sub-task or the combined task. In particular, we will introduce object placement in Section~\ref{sec:object_placement}, image blending in Section~\ref{sec:image_blending}, image harmonization in  Section~\ref{sec:image_harmonization}, shadow (\emph{resp.}, reflection) generation in Section~\ref{sec:shadow_generation} (\emph{resp.}, \ref{sec:reflection_generation}), generative image composition in Section~\ref{sec:generative_composition}, foreground object search in Section~\ref{sec:foreground_object_search}.  
In each section, we will introduce the existing methods, available datasets, and common evaluation metrics. 
Finally, we will conclude the whole paper in Section~\ref{sec:conclusion}. The contributions of this paper can be summarized as follows,
\begin{itemize}
    \item To the best of our knowledge, this is the first comprehensive survey on deep image composition (object insertion).
    \item We summarize the issues in image composition as three types of inconsistencies. We clarify the relation between inconsistency, issue, sub-task, and pipeline in Fig.~\ref{fig:summary}. We also summarize the issues that previous works attempt to solve in Table \ref{tab:method_summary}. All the above summaries give rise to a large picture for deep image composition. 
    \item For each sub-task and the combined task, we survey the existing methods, available datasets, and common evaluation metrics. We also show some experimental results. We believe that this comprehensive survey can serve as the roadmap for the future research in a broad community. 
\end{itemize}

%% file: sections/object_placement.tex
\section{Object Placement}\label{sec:object_placement}

\begin{figure*}[t]
\begin{center}
\includegraphics[width=.9\linewidth]{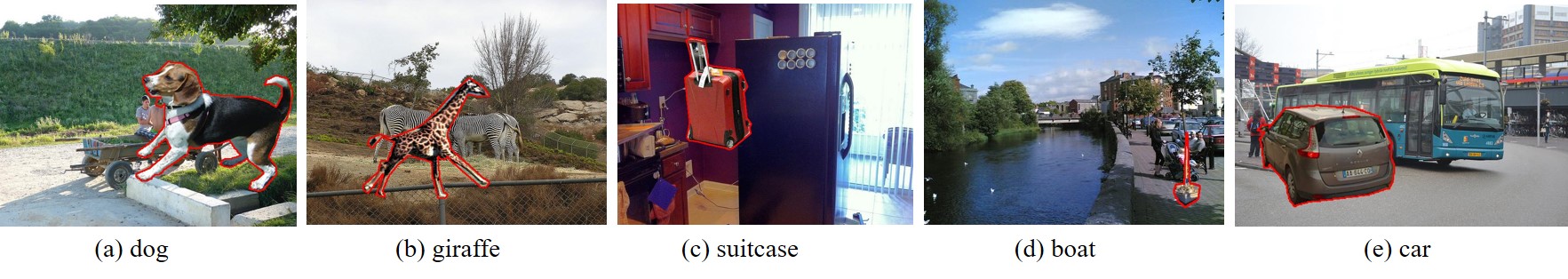}
\end{center}
\caption{Examples of unreasonable object placements. The inserted foreground objects are marked with red outlines. From left to right: (a) objects with inappropriate size; (b) unreasonable occlusion; (c) objects hanging in the air; (d) objects appearing at the semantically unreasonable place;  (e) inconsistent perspectives.
}
\label{fig:object_placement}
\end{figure*}

\begin{figure*}[t]
\begin{center}
\includegraphics[width=.9\linewidth]{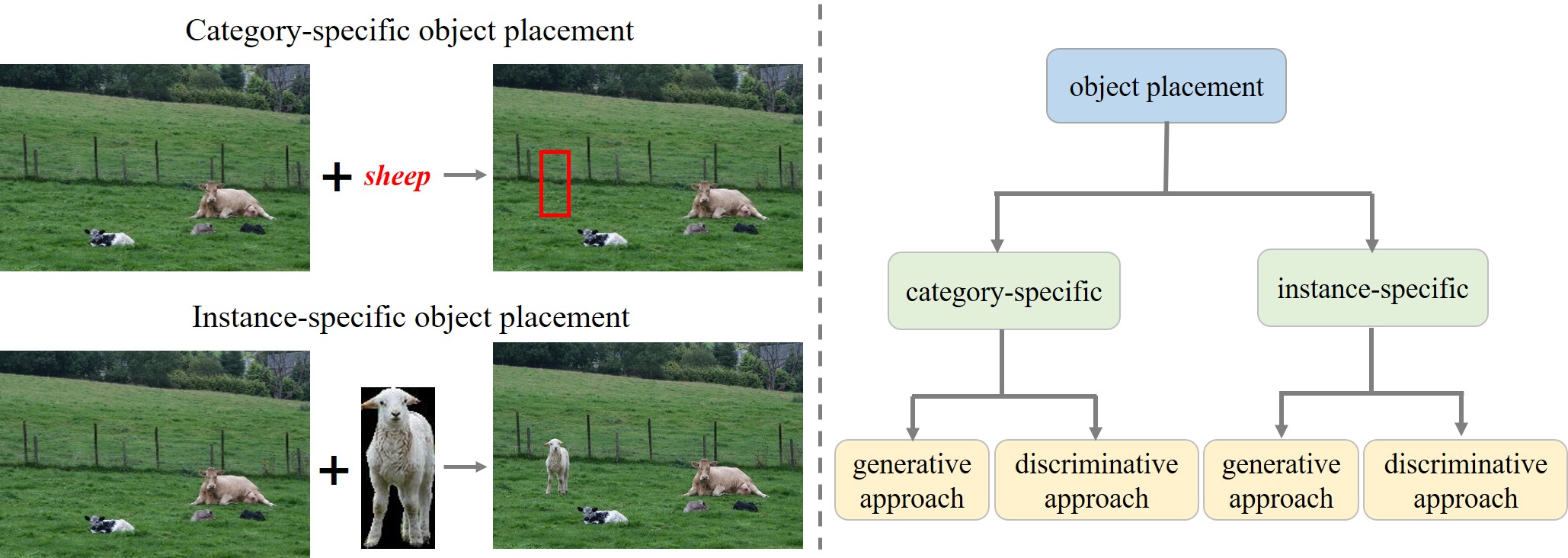}
\end{center}
\caption{In the left subfigure, we compare category-specific object placement with instance-specific object placement. In the right subfigure, we show the taxonomy of existing object placement methods. 
}
\label{fig:object_placement_taxonomy}
\end{figure*}

Object placement aims to paste the foreground on the background with suitable location, size, and shape. As shown in Fig.~\ref{fig:object_placement}, the cases of unreasonable object placement  include but are not limited to: a) the foreground object has inappropriate size (\emph{e.g.}, the dog is too large); b) the foreground object has unreasonable occlusion with background objects (\emph{e.g.},  the fences are unreasonably occluded by the giraffe); c)  the foreground object does not have reasonable force condition (\emph{e.g.}, the suitcase is floating in the air); d) the foreground object appears at a semantically unreasonable place (\emph{e.g.}, the boat appears on the land); e) inconsistent perspectives between foreground and background (\emph{e.g.}, the car and the bus have inconsistent perspectives).
By taking all the above factors into consideration, object placement is a very challenging task.

\subsection{Traditional Methods}
Some object placement methods designed explicit rules to find reasonable location and scale for the foreground object.
For example, \citet{LSCPRemez2018} proposed to move the foreground object of fixed scale along the same horizontal scanline on the background. They assumed that the locations along the same horizontal scanline have similar depth, so that the true scale of foreground object can be well-preserved. 
\citet{InstanceSwitchingWang2019} designed the instance-switching strategy to generate new  images through switching different instances of the same class with similar shape and scale. 
To better refine the position where the object is pasted, \citet{InstaBoostFang2019} explored appearance consistency heatmap to guide the object placement, based on the intuition that an object could be moved to another location which has similar visual context to its original visual context. 
Specifically, one element in the appearance consistency heatmap measures the similarity between the visual context at this point and original visual context. \citet{STDODISGeorgakis2017} proposed to combine support surface detection and semantic segmentation to find proper location for placing the object. With the determined location, the size of the object is decided in the light of the depth at this location and the original scale of the object. \citet{WhatWhereZhang2020} proposed to model the probability distribution of bounding box information conditioned on background image and foreground category using Gaussian mixture model. 

Although these rules are effective in some cases,  they are incomplete and sometimes inaccurate, which is far below the requirement to handle the diverse and complicated challenges in object placement task.

\begin{figure*}[t]
\begin{center}
\includegraphics[width=.92\linewidth]{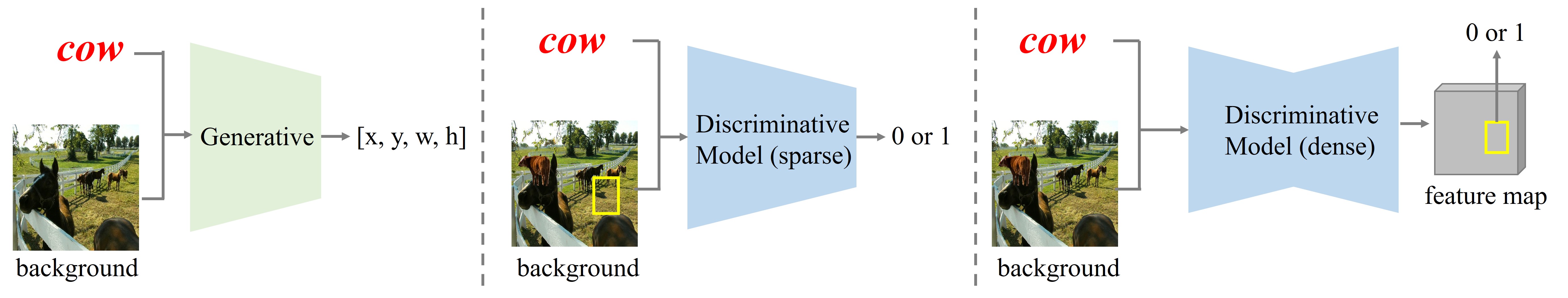}
\end{center}
\caption{We show three types of methods for category-specific object placement. \emph{Generative model}: given the foreground category and background image, the model generates a reasonable bounding box (\emph{e.g.}, location (x,y) and scale (w,h)). \emph{Sparse discriminative model}: given the foreground category, foreground bounding box, and background image, the model predicts a rationality score.  \emph{Dense discriminative model}: given the foreground category and background image, the model uses sliding window on the feature map to get the rationality score for each bounding box. 
}
\label{fig:object_placement_three_types_a}
\end{figure*}

\begin{figure*}[t]
\begin{center}
\includegraphics[width=.95\linewidth]{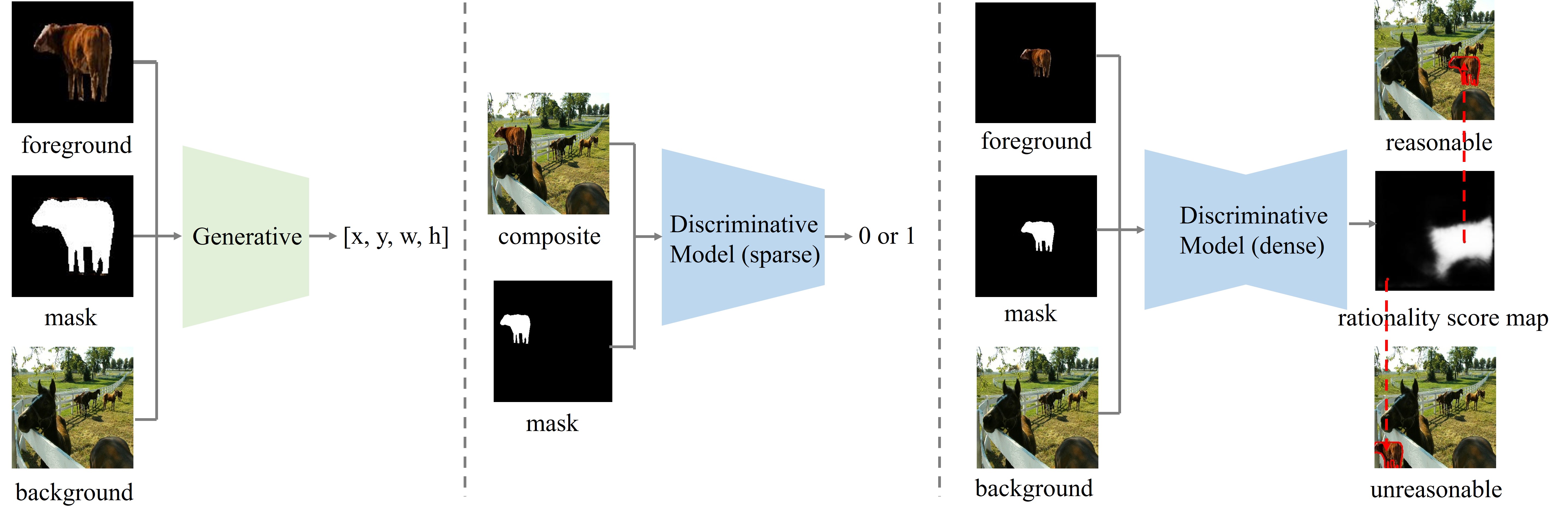}
\end{center}
\caption{We show three types of methods for instance-specific object placement.
\emph{Generative model}: given the foreground, foreground object mask, and background, the model generates a reasonable placement (\emph{e.g.}, location (x,y) and scale (w,h)) for the foreground. \emph{Sparse discriminative model}: given the composite image and composite foreground mask, the model predicts a rationality score.  \emph{Dense discriminative model}: given the foreground, foreground object mask, and background, the model predicts a rationality score map containing the rationality scores for all locations. 
}
\label{fig:object_placement_three_types_b}
\end{figure*}

\begin{figure*}[t]
\begin{center}
\includegraphics[width=.9\linewidth]{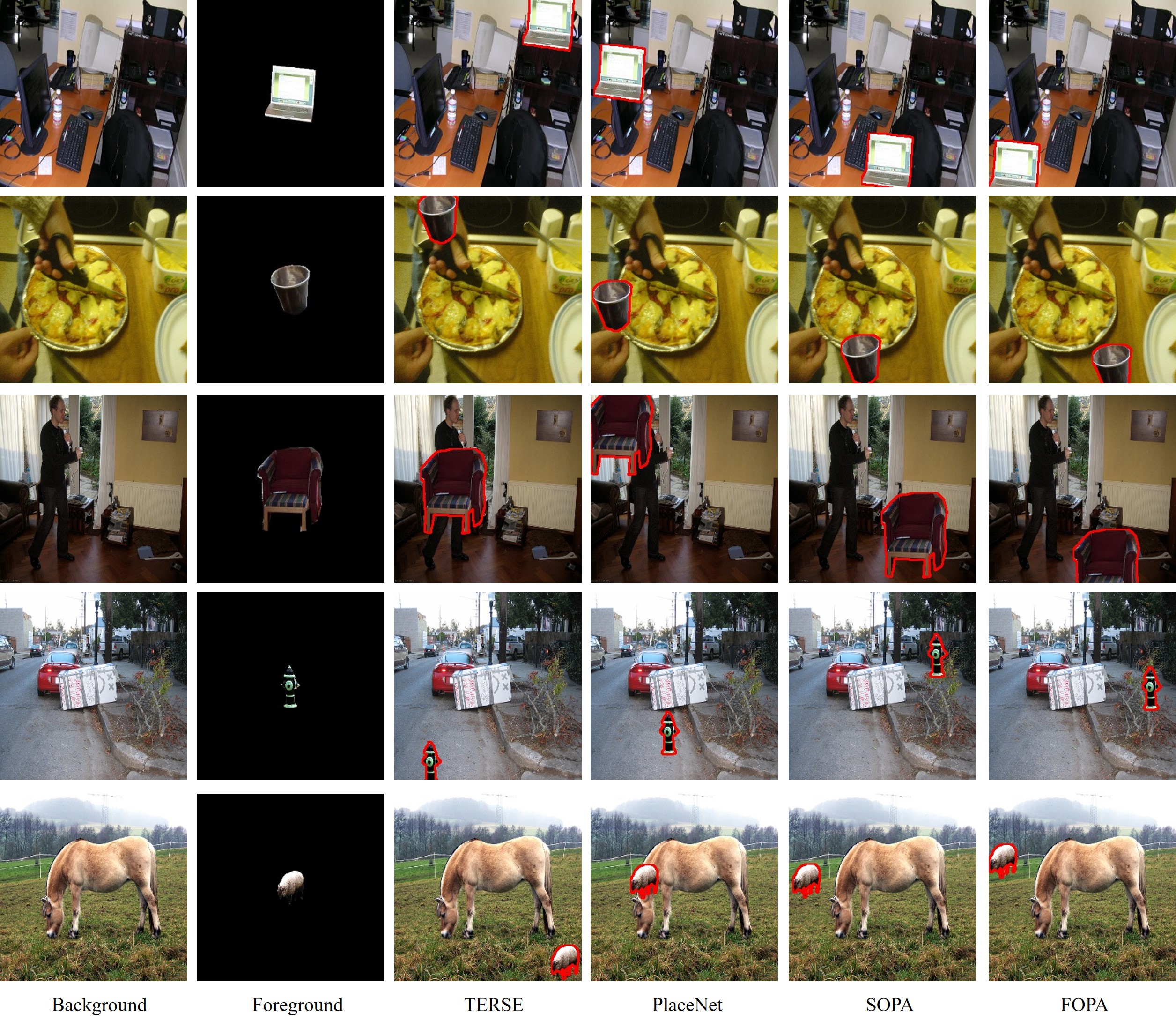}
\end{center}
\caption{The visualization results of different object placement methods on OPA \cite{liu2021opa} dataset. From left to right in each row, we show the foreground, background, and composite images obtained by TERSE \cite{SyntheticTripathi2019}, PlaceNet \cite{LearningObjPlaZhang2020}, SimOPA (SOPA) \cite{liu2021opa}, FOPA \cite{niu2022fast}. 
}
\label{fig:object_placement_results}
\end{figure*}

\subsection{Deep Learning Methods}
Apart from the above methods which design explicit rules to infer the reasonable placement for the foreground object, some methods~\cite{zhan2019spatial,CompositionalGANAzadi2020,zhanhierarchy,tan2019image,wang2017binge} employ deep learning techniques to predict the placement and generate the composite image automatically.

The existing deep learning based object placement methods can be divided into category-specific object placement and instance-specific object placement. For category-specific object placement, the model aims to predict plausible bounding boxes given a background image and a foreground category. This group of methods assume that the predicted bounding boxes are suitable for all instances belonging to the same foreground category. Nevertheless, this assumption is too restrictive, because different instances from the same category may have distinct properties (\emph{e.g.}, geometry, fine-grained semantics) and thus require bounding boxes with different scales/locations. In contrast, instance-specific object placement methods aim to predict plausible spatial transformations given a background image and a specific foreground object. The difference between category-specific object placement and instance-specific object placement is shown in Fig. \ref{fig:object_placement_taxonomy}. Next, we will introduce these two groups of works separately. 

\subsubsection{Category-specific Object Placement}

Category-specific object placement methods can be categorized into generative approach and discriminative approach. The generative approach targets at predicting one or multiple reasonable bounding boxes for the foreground category, whereas the discriminative approach aims to predict the rationality score of a bounding box for certain foreground category. The discriminative approach can be further divided into sparse discriminative approach and dense discriminative approach. The sparse discriminative approach takes in a background image with foreground bounding box and predicts a rationality score. The dense discriminative approach takes in a background and produces a feature map, based on which sliding window is used to predict the rationality score of each bounding box. The comparison between generative model, sparse discriminative model, and dense discriminative model is illustrated in Fig. \ref{fig:object_placement_three_types_a}. 

\textbf{Generative approaches:} 
\citet{Tan2018WhereAW} proposed to predict the location and scale of inserted object by taking the background image and object layout as input. Besides, the bounding box prediction task is converted to a classification task by discretizing the locations and scales.  
\citet{ContextawareLee2018} investigated on taking a background semantic map instead of a background image as input.  Given a background semantic map, they designed a network consisting of two generative modules, in which the first module accounts for the bounding box of inserted object and the second module accounts for the mask shape of inserted object.  \citet{parihar2024text2place} focused on ``person" category and leveraged the prior knowledge of text-to-image generation model. Specifically, the foreground mask and foreground image are jointly optimized to fit the background and text prompt, after which the foreground mask can be used to indicate the person placement.   

\textbf{Discriminative approaches:} The methods in \cite{MVCIKAODDDvornik2018,dvornik2019importance} used a network to predict whether a bounding
box is suitable for certain foreground category, based on the contextual information surrounding the bounding box. This approach needs to pass through the network once for each bounding box, which is very time-consuming. To accelerate this process,   \citet{volokitin2020efficiently} employed masked convolutions to aggregate the contextual information along four directions as context feature maps, based on which the contextual information excluding each bounding box can be obtained efficiently to predict the rationality score for this bounding box. 

\subsubsection{Instance-specific Object Placement}

Instance-specific object placement methods can also be categorized into generative approach and discriminative approach. The generative approach targets at predicting one or multiple reasonable placements (\emph{i.e.}, spatial transformations) for the foreground object, whereas the discriminative approach aims to predict the rationality score of a composite image in terms of the foreground object placement. The discriminative approach can be further divided into sparse discriminative approach and dense discriminative approach. The sparse discriminative approach takes in a composite image and predicts a rationality score. The dense discriminative approach takes in a pair of foreground and background, and predicts a rationality score map. The comparison between generative model, sparse discriminative model, and dense discriminative model is illustrated in Fig. \ref{fig:object_placement_three_types_b}. 

\textbf{Generative approaches:} Generative approaches \cite{SyntheticTripathi2019,LearningObjPlaZhang2020,STGANChen2018,li2021image,li2019advertisement,
ye2023efficient,wang2024learning,cheng2025diverse} predict different types of spatial transformations (\emph{e.g.}, shifting and scaling, affine transformation, perspective transformation) for the foreground object, which is more flexible and powerful than category-specific object placement methods.
For instance, \citet{SyntheticTripathi2019} developed a model with generator, discriminator, and target network. Given a pair of background and foreground, the generator predicts an affine transformation for the foreground object to produce a composite image. The produced composite image is expected to fool the discriminator and fit the target network corresponding to a downstream task (\emph{e.g.}, object detection). 
To produce multiple reasonable placements, \citet{LearningObjPlaZhang2020} combined the foreground feature, background feature, and a random vector to predict the object placement. Moreover, they ensured the diversity of object placement by enforcing the pairwise distances between predicted placements to approach those between corresponding random vectors. 
To promote the diversity of generated placements, 
\citet{graco2022} established the bijection between random vector and positive composite image. 
Moreover, they reformulated object placement as a graph completion task. In particular, background nodes have both content features and placements, while the inserted foreground node only has content feature, giving rise to an incomplete graph. Hence, they estimated the missing placement for the foreground node to complete the graph.
\citet{oprl2023} proposed to make sequential decisions to produce a reasonable placement by using reinforcement learning. \citet{qin2025think} employed a pre-trained large multi-modal model to generate a caption containing the placement information, and then predicted the placement bounding boxes. \citet{zhou2025bootplace} first predicted candidate bounding boxes and then associated each object query with matched bounding box. 

In terms of more advanced geometric transformation like view synthesis and pose transfer, some methods \cite{CompositionalGANAzadi2020,STGANChen2018,strat,gou2023virtual} predicted perspective transformation to adjust the viewpoint and some methods \cite{wang2017binge,yao2023scene} predicted human pose in the scene context. 
\citet{zhan2019spatial} adopted spatial transformer network (STN) \cite{STNJaderberg2015} to predict the warping parameters under an adversarial learning framework. 
\citet{CompositionalGANAzadi2020} employed STN to warp the foreground and relative appearance flow network to change the viewpoint of foreground. Additionally, they investigated on self-consistency constraint, that is, the generated composite image could be decomposed back to the foreground and background. 
ST-GAN \cite{STGANChen2018} proposed to warp a foreground object into a background image with iterative spatial transformations predicted by STN. As a follow-up work, \citet{kikuchi2019regularized} replaced the iterative spatial transformations in \cite{STGANChen2018} with one-shot spatial transformation. 
\citet{gou2023virtual} revealed that for perspective transformation, predicting the target locations of four source points is more effective than predicting the locations for more source points \cite{strat} or predicting the transformation parameters \cite{STGANChen2018}.
However, the view and pose synthesis ability is quite limited, especially for drastic viewpoint change (\emph{e.g.}, from front view to side view) and complicated human-object interaction (\emph{e.g.}, playing piano). To accomplish drastic viewpoint change and flexible pose transfer, generative composition methods attempted to re-generate the foreground object, which will be introduced in Section~\ref{sec:generative_composition}.

\textbf{Discriminative approaches:} \citet{liu2021opa} proposed a discriminative approach named SimOPA to verify whether a composite image is rational in terms of the foreground object placement. Particularly, they feed the concatenation of composite image and foreground mask into a binary classification network to predict a rationality score. However, this discriminative approach is very inefficient, because they need to go through the discriminative network multiple times to find a reasonable object placement. To address this issue, \citet{niu2022fast} dubbed SimOPA as slow object placement assessment (SOPA) model and proposed a fast object placement assessment (FOPA) model, which can predict the rationality scores at all locations by going through the model only once. Precisely, they take in a pair of background and scaled foreground, and produce a rationality score map, in which each entry represents the rationality score of the composite image obtained by pasting the foreground at this location. They developed several innovations (\emph{e.g.}, background prior transfer, feature mimicking) to bridge the performance gap between FOPA and SOPA, reaching the conclusion that FOPA can achieve comparable performance with SimOPA at significantly reduced cost. FOPA~\cite{niu2022fast} has also demonstrated stronger ability to generate realistic composite images than generative approaches \cite{graco2022,oprl2023}. 
Similar to FOPA~\cite{niu2022fast}, \citet{topnet2023,poska2025hopnet} also proposed to predict the rationality scores of all scales and locations, based on the interaction output between foreground and background. \citet{gao2025object} explored extending object placement task to a broader range of foreground categories and background scenes. 

In the end, we briefly discuss the occlusion issue. Most of the above methods seek for reasonable placements to avoid the occurrence of occlusion, \emph{i.e.}, the inserted foreground is not occluded by background objects. Differently, a few methods~\cite{CompositionalGANAzadi2020,zhanhierarchy,tan2019image,ghoneim2024depgan} attempt to address the unreasonable occlusion when it occurs. Specifically, they first estimate the relative depth relation between the foreground object and the surrounding background objects. Then, they remove the occluded part of foreground object. In this way, they are able to generate composite images with reasonable inter-object occlusions.

\subsection{Datasets and Evaluation Metrics}

In some early works \cite{SyntheticTripathi2019,InstaBoostFang2019}, object placement is used as data augmentation strategy to facilitate the downstream tasks (\emph{e.g.}, object detection, instance segmentation). Therefore, they make use of existing object detection and instance segmentation datasets~\cite{lin2014microsoft,PascalVOCMark2010,CityscapesMarius2016,GMUKitchenGeorgios2016}. In particular, the foregrounds are cropped out based on the annotated segmentation masks. After removing the foreground objects, the remaining incomplete background images are restored to complete background images by using image inpainting techniques~\cite{yeh2017semantic,liu2018image,yu2019free}. In this manner, triplets of foregrounds, backgrounds, and ground-truth composite images can be obtained.  
Some other works focus on specific applications like 2D virtual try-on~\cite{STGANChen2018,kikuchi2019regularized,li2021image}  (\emph{e.g.}, placing glasses/hats on human faces) or logo composition~\cite{li2019advertisement} (\emph{e.g.}, attaching logo to product image), so they need to collect foregrounds and backgrounds specifically for these applications. More recently, \citet{liu2021opa} released a large-scale object placement assessment dataset named \emph{OPA}, which consists of 73,470 composite images and their binary rationality labels. OPA dataset is constructed by compositing the foregrounds and backgrounds from COCO dataset \cite{lin2014microsoft}, followed by manually labelling the rationality of obtained composite images. A large number of annotated composite images could greatly facilitate the research on object placement. \citet{qin2025think} established the OPAZ dataset following the format of OPA. 

To evaluate the quality of generated composite images, previous object placement works usually adopt the following schemes: 1) Some works measure the similarity between real images and composite images. For example, \citet{Tan2018WhereAW} score the correlation between the distributions of predicted boxes and ground-truth boxes. \citet{LearningObjPlaZhang2020} calculate Frechet Inception Distance (FID) \cite{FIDHeusel2017} between composite images and real images. However, they cannot evaluate each individual composite image.
2) Some works \cite{SyntheticTripathi2019, InstaBoostFang2019} utilize the performance improvement of downstream tasks (\emph{e.g.}, object detection) to evaluate the quality of composite images, where the training sets of the downstream tasks are augmented with generated composite images. However, the evaluation cost is quite huge and the improvement in downstream tasks may not reliably reflect the quality of composite images, because it has been revealed in \cite{SimpleCopyPasteGhiasi2020} that randomly generated unrealistic composite images could also boost the performance of downstream tasks. 
3) Another common evaluation strategy is user study, where people are asked to score the rationality of object placement \cite{ContextawareLee2018, Tan2018WhereAW}. User study complies with human perception and each composite image can be evaluated individually.  OPA dataset~\cite{liu2021opa} has annotated composite images and its test set could be used for evaluation. Nevertheless, the sparse annotations only cover a small proportion of locations and scales, which limits the universal evaluation of arbitrary composition results. 4) To support the  evaluation of arbitrary composition results, we can train a binary classifier based on annotated positive and negative composite images, to predict the rationality score of an arbitrary composite image. 

\subsection{Experiments}

In this section, we focus on instance-specific object placement and compare existing object placement methods for generating a reasonable composite image. For ease of comparison, we fix the foreground scale and only predict the reasonable location for the foreground object. 
Recall that instance-specific object placement methods are divided into generative approaches and discriminative approaches.
For generative approach, we choose TERSE \cite{SyntheticTripathi2019} and PlaceNet \cite{LearningObjPlaZhang2020}, which can directly predict one placement. For discriminative approach, we report the results of SimOPA \cite{liu2021opa} and FOPA \cite{niu2022fast}. We use SimOPA and FOPA to generate rationality score map, based on which the location with the largest rationality score is chosen as the optimal placement. We train and evaluate different methods on OPA dataset \cite{liu2021opa}. The test results are shown in Fig.~\ref{fig:object_placement_results}, from which it can be seen that discriminative approaches usually achieve better results than generative approaches. One possible explanation is that TERSE \cite{SyntheticTripathi2019} and PlaceNet \cite{LearningObjPlaZhang2020} only utilize the annotated composite images to update the discriminator, without fully using the annotations to train the generator. Nevertheless, discriminative approaches also have failure cases when dealing with occlusion and complex scenes (\emph{e.g.}, unreasonable occlusion between fire hydrant and fallen branches in row 4).

For practical usage, we recommend dense discriminative approaches  \cite{niu2022fast,topnet2023}, which are more stable, effective, and flexible than generative approaches and sparse discriminative approaches. Dense discriminative approaches can efficiently identify reasonable bounding boxes (\emph{i.e.}, location, scale) for the inserted foreground object. Then, if necessary, generative composition methods could be used to further adjust the viewpoint and pose of foreground object (see Section~\ref{sec:generative_composition}).

%% file: sections/image_blending.tex
\section{Image Blending}\label{sec:image_blending}

\begin{figure*}[t]
\begin{center}
\includegraphics[width=.93\linewidth]{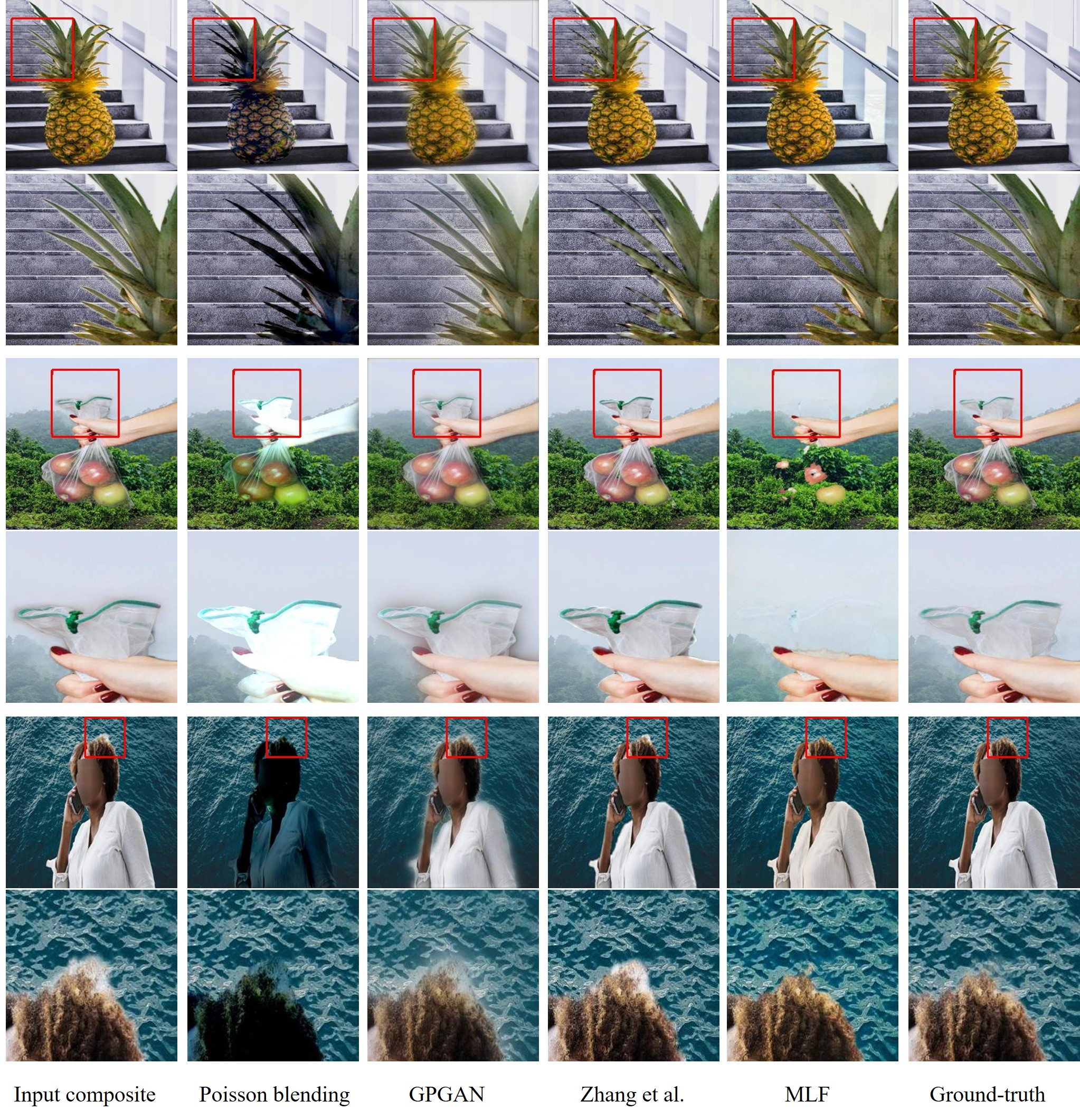}
\end{center}
\caption{The leftmost column is the initial composite image obtained using the alpha matte predicted by LFPNet~\cite{liu2021long}. The rightmost column is the ground-truth composite image obtained using ground-truth alpha matte. The middle columns are the refined results obtained by Poisson image blending \cite{perez2003poisson}, GP-GAN \cite{wu2019gp}, \citet{zhang2020deep}, and 
MLF \cite{zhang2021deep}. The odd rows display the whole image, while the even rows zoom in the red bounding boxes in the odd rows for better observation. 
}
\label{fig:image_blending_results}
\end{figure*}

During image composition, the foreground is usually extracted using image segmentation \cite{minaee2021image} or matting \cite{xu2017deep} methods. However, the segmentation or matting results may be noisy and the foregrounds are not precisely delineated. When the foreground with jagged  boundaries is pasted on the background, there will be abrupt intensity change between the foreground and background. To refine the boundary and reduce the fuzziness,  image blending techniques have been developed. 

\subsection{Traditional Methods}

Traditional image blending methods aim to smooth the transition between foreground and background. Alpha blending \cite{porter1984compositing} proposed to assign alpha values for boundary pixels indicating what fraction of the colors are from foreground or background, in which the alpha values need to be manually set. Alpha blending is a simple and fast method, but it blurs the fine details and brings in ghost effects. 
Considering multi-scale information, Laplacian pyramid  blending \cite{burt1983multiresolution} proposed to build multi-scale Laplacian pyramids for two images and perform alpha blending at each scale. Then, the final output is obtained by adding up the blended results of different scales. 

Another group of methods attempt to achieve smooth
boundary transition by enforcing gradient domain smoothness \cite{fattal2002gradient,kazhdan2008streaming,levin2004seamless,szeliski2011fast}. The earliest work along this research direction is Poisson image blending \cite{perez2003poisson}. Poisson image blending \cite{perez2003poisson} proposed to enforce the gradient domain consistency with respect to the source image containing the foreground, where the gradient of inserted foreground is computed and propagated from the boundary pixels in the background. Although Poisson image blending can yield more pleasant results than simple alpha blending, it is very time-consuming to solve the Poisson equation. Therefore, there are many follow-up works \cite{szeliski2011fast,tanaka2012seamless,kazhdan2008streaming} to accelerate Poisson image blending by using different techniques. Based on the observation that the effectiveness of Poisson image blending seriously depends on the boundary condition, \cite{jia2006drag} designed a method to optimize the boundary condition. 
To avoid the color bleeding and halo effect brought by Poisson image blending, \citet{tao2010error} developed a two-step algorithm: first processing the gradient values on the boundary and then employing a weighted integration scheme to reconstruct the image from its gradient field. The above methods based on gradient domain smoothness can smooth the transition between foreground and background to some extent. However, background colors may seep through the foreground too much and distort the foreground color, which would bring significant loss to the foreground content. 

\subsection{Deep Learning Methods}

Inspired by traditional image blending methods \cite{perez2003poisson,burt1983multiresolution}, some recent works \cite{wu2019gp,zhang2020deep} explored incorporating the function of smoothing boundary into deep learning network.  
Among them,  the works \cite{wu2019gp,zhang2020deep,zhang2021deep} not only enabled smooth transition over the boundary, but also reduced the illumination discrepancy between foreground and background, in which the latter one is the goal of image harmonization in Section~\ref{sec:image_harmonization}. In this section, we only introduce the way they enable smooth transition over the boundary. These two works \cite{wu2019gp,zhang2020deep} are both inspired by \cite{perez2003poisson}. Specifically, they added the gradient domain constraint to the objective function according to Poisson equation, which can produce a smooth blending boundary with gradient domain consistency. They both optimized over the input composite image to minimize the gradient domain loss. Differently, \cite{wu2019gp} had a close-form solution, while \cite{zhang2020deep} converted the gradient domain loss to a differentiable loss function and uses gradient descent algorithm. 

Different from \cite{wu2019gp,zhang2020deep} which are inspired by traditional image blending, recent works \cite{zhang2021deep,Xing0WZC22} proposed learnable image blending, which produces a seamlessly blended image by taking in a pair of foreground image and background image.
Specifically, the fusion network in \citet{zhang2021deep} used two separate encoders to extract and fuse multi-scale features from foreground and background. Because the fusion network relies on ground-truth composite images obtained by using accurate alpha matte as supervision, the work \cite{zhang2021deep} also proposed an easy-to-hard data-augmentation scheme to relieve the burden of annotating ground-truth alpha matte. Similarly, \citet{Xing0WZC22} proposed to concatenate foreground image, background image, and imperfect mask as input to generate a blended image. For these methods \cite{zhang2021deep,Xing0WZC22}, the trained models have the ability to refine imperfect masks and deliver more naturally blended images. Besides the GAN-based generative models~\cite{zhang2021deep,Xing0WZC22}, conditional diffusion models~\cite{zhang2023adding,labs2025flux} could also generate blended image conditioned on the composite image, demonstrating stronger generalization ability. For the above methods, the quality of initial mask has significant impact on the quality of blended image. If the initial mask is of very poor quality, these methods can hardly produce high-quality blended image. 

More recently, mask-free image blending has emerged, which does not require initial masks. ControlCom~\cite{zhang2023controlcom} received a foreground image enclosing the foreground object and a background image with bounding box specifying the foreground placement, producing a composite image. The mask-free methods relieve the burden of initial mask prediction, which is not affected by the quality of initial masks. In \cite{zhang2023controlcom}, the foreground object features are injected into diffusion model via cross-attention. Such strategy to inject foreground information may lead to the slight loss or distortion of foreground details. In contrast, the in-context learning strategy to inject foreground information~\cite{song2025insert}, that is, concatenating foreground image as one of the inputs, can better preserve the foreground details.

\subsection{Datasets and Evaluation Metrics}
To the best of our knowledge, there are only few deep learning methods \cite{wu2019gp,zhang2020deep,zhang2021deep} for image blending and there is no unified benchmark dataset. 
\citet{zhang2020deep} do not mention the source of used images. \citet{wu2019gp} manually crop objects from transient attributes database \cite{laffont2014transient} to create input composite images. Similarly, \citet{zhang2021deep} take foreground images from segmentation datasets \cite{hou2017deeply,shen2016automatic} and random background images to construct input pairs. 

The existing deep image blending works~\cite{wu2019gp,zhang2020deep,zhang2021deep} adopt the following evaluation metrics: 1) \citet{zhang2021deep} deem the composite images obtained using ground-truth alpha matte as ground-truth composite image, and calculate Peak Signal-to-Noise Ratio (PSNR) between resultant image and ground-truth composite image.  2) conducting user study by asking engaged users to select the most realistic images; 
3) calculating realism score using the pretrained model~\cite{zhu2015learning} which reflects the realism of a composite image.

\subsection{Experiments}

We evaluate different image blending methods conditioned on the matting results. First, we create composite images using the alpha mattes predicted by the state-of-the-art trimap-based image matting methods \cite{dai2022boosting,liu2021tripartite,liu2021long}. Then, we hope that image blending methods can refine the obtained composite images. 
We sample 500 foreground images from recent image matting datasets \cite{liu2021long,li2021deep,li2021privacy}. For each foreground image, we randomly select two background images from BG20K~\cite{li2022bridging}. The foreground images and background images form the test set. 

By taking LFPNet~\cite{liu2021long} as an example matting method, we predict the alpha mattes and obtain the composite images. We observe that LFPNet can generally achieve satisfactory results except some challenging cases. We select several of its failure cases to verify the effectiveness of image blending methods. 

We report the results of Poisson image blending \cite{perez2003poisson}, GP-GAN \cite{wu2019gp}, \citet{zhang2020deep}, and MLF \cite{zhang2021deep}. We also report the ground-truth composite image obtained using ground-truth alpha matte for comparison. From Fig. \ref{fig:image_blending_results}, it can be seen that the composite images obtained using predicted alpha mattes are very close to the ground-truth composite image except partial boundary regions. We observe that Poisson image blending \cite{perez2003poisson} smooths the transition boundary to some extent, but unexpectedly distorts the foreground content by seeping through the foreground. GP-GAN \cite{wu2019gp} and \citet{zhang2020deep} are inspired by Poisson image blending, but use content loss to preserve the original foreground content. Therefore, they strike a balance between preserving the foreground content and smoothing the boundary. However, some smoothed boundary regions are still not satisfactory. MLF \cite{zhang2021deep} can obtain visually appealing results in some
cases. Nonetheless, it may erase detailed information (\emph{e.g.}, the small leaves of a pineapple) and fail in handling transparent
foreground objects (\emph{e.g.}, plastic bag).

For practical usage, traditional image blending methods are recommended under very specific circumstances. For example, when the foreground mask is accurate and there is no high demand for sharp boundary, alpha blending \cite{porter1984compositing} is applicable. When the colors of foreground boundaries are expected to be close to the colors of adjacent background boundaries, Poisson blending  \cite{perez2003poisson} is applicable. Otherwise, conditional diffusion models \cite{labs2025flux, song2025insert} are recommended. When high-quality initial mask is available, diffusion model \cite{labs2025flux} conditioned on composite image is recommended. When initial mask is unavailable or only poor-quality initial mask is available, diffusion model \cite{song2025insert} conditioned on background image and foreground image is recommended.

%% file: sections/image_harmonization.tex
\section{Image Harmonization}\label{sec:image_harmonization}

Given a composite image, its foreground and background are likely to be captured in different conditions (\emph{e.g.}, weather, season, time of day, camera setting), and thus have distinctive illumination characteristics, which make them look incompatible. Image harmonization aims to adjust the appearance of composite foreground according to composite background to make it compatible with the composite background. We classify the existing methods into rendering based and non-rendering based methods according to whether using rendering techniques. 

\subsection{Rendering based Methods}

Conventional image relighting \cite{pandey2021total,shu2017portrait,wang2020single,zhang2021neural,tewari2021monocular}
aims to adjust the appearance of an image or the object in an image as lit by novel illumination. 
With some adaptation, image relighting  can also be used to adjust the foreground appearance according to the illumination of a new background \cite{pandey2021total,zhang2021neural,careaga2023intrinsic,hu2024spatially}, which bears some resemblance to image harmonization. However, they usually achieve this goal by inferring explicit illumination condition, material properties, or 3D geometry, in which the supervision for these information is difficult and expensive to acquire. Besides, they generally have strong assumption about the light source, which may not generalize well to complicated real-world scenes. 

\subsection{Non-rendering based Methods}

Early traditional image harmonization methods \cite{xue2012understanding,multi-scale,lalonde2007using,song2020illumination} performed color transformation on the foreground to match the low-level color statistics between foreground and background. The difference between different methods mainly lies in the matching details. 
For example, \citet{xue2012understanding} predicted the histogram zone (\emph{e.g.}, low, middle, high)  which can be best matched between foreground and background, and then adjusted the foreground color to match the selected zone between foreground and background. \cite{multi-scale} explored decomposing an image into a multi-resolution pyramid with  multiple subbands, and performing histogram matching for each subband between foreground and background. 
\citet{lalonde2007using} proposed to represent foreground and background with color clusters, followed by matching foreground and background color clusters. 
\citet{song2020illumination} proposed to calculate the color transformation (channel-wise scales) based on the gray pixels of foreground and background, because normalized illumination color can be directly derived from the pixel values of gray pixels. In some works on sky replacement \cite{zou2020castle,tao2009skyfinder,tsai2016sky}, they attempted to match the color statistics (\emph{e.g.}, mean, variance) between sky region and non-sky region. Broadly speaking, traditional color transfer methods \cite{reinhard2001color,xiao2006color,fecker2008histogram,pitie2007automated,faridul2014survey,alappatt2016survey} can also be used for color matching between foreground and background to produce a harmonized image. 

Early deep learning based image harmonization methods target at making the harmonized images indistinguishable from real images. For instance, \citet{zhu2015learning} explored predicting the realism of an image using a CNN classifier. With such realism predictor, they learned the color transformation for the foreground to achieve high realism score, and also enforced the color variation in different channels to be close. 
Similar to \cite{zhu2015learning}, 
the works \cite{zhan2020adversarial,chen2019toward} used adversarial learning to make the harmonized images indistinguishable from real images.
\citet{bhattad2020cutandpaste} drew inspiration from Retinex theory \cite{kimmel2003variational} that an image can be decomposed into albedo (reflectance) and shading (illumination). On the premise of this assumption, an image harmonization model is trained so that the harmonized result should have consistent albedo and consistent background shading with input composite image.

With the emergence of image harmonization datasets consisting of paired training data (see Section \ref{sec:image_harmonization_dataset}), abundant image harmonization methods \cite{tsai2017deep,DoveNet2020,cong2022high,LEMaRT,CaiTMM2022,
DucoNet,ChenArXiv2023,yu2023semantic,wang2023harmonized,peng2024frih,wang2024retrieval,li2025image,zhang2025region} using paired supervision have been developed. 
\citet{tsai2017deep} proposed the first end-to-end CNN network for image harmonization and leveraged auxiliary semantic segmentation branch to enhance the basic image harmonization network. Another work \citet{sofiiuk2021foreground} also utilized high-level semantic features, which are inserted into the encoder to provide auxiliary information. 
\citet{xiaodong2019improving} designed an additional Spatial-Separated Attention Module to deal with foreground and background feature maps separately.
\citet{Hao2020bmcv} employed self-attention \cite{wang2018non} mechanism  to propagate relevant information from background to foreground. 
By treating different capture conditions as different domains, \citet{DoveNet2020} proposed a domain verification discriminator to pull close the foreground domain and background domain. Similarly, \citet{bargain} formulated image harmonization as background-guided domain translation task, in which the domain code of background is directly used to guide the harmonization process. One byproduct of \cite{bargain} is predicting the inharmony level of an image by comparing the domain codes of foreground and background, so that we can selectively harmonize those inharmonious composite images. Inspired by \cite{bargain}, \citet{ValanarasuICLR2023} proposed to extract style code from part of background. 

In \cite{regionaware}, they reframed image harmonization as a background-to-foreground style transfer problem and introduced region-aware adaptive instance normalization (AdaIn) to transfer visual style from the background to the foreground. A succeeding work \cite{hang2022scs} extended \cite{regionaware} by searching foreground-relevant background regions and transferring foreground-relevant style from background to foreground. They also extended the triplet loss in \cite{bargain} to contrastive loss. Some subsequent works \cite{ju2023adaptive,chen2024segment} adopted the similar idea of searching the background regions matching the foreground region. 

Analogous to \cite{bhattad2020cutandpaste}  using Retinex theory, \citet{intriharm} also developed a model to disentangle a composite image into reflectance map and illumination map, in which the illumination map is harmonized by transferring lighting information from background to foreground. Another work \cite{guo2021image} also adopted the similar decomposition network and explored integrating transformer block \cite{vaswani2017attention} into the network, which is further extended to \cite{GuoPAMI2022}. Following the disentanglement technical route, \citet{jiang2021ssh} proposed to disentangle an image into content representation and appearance representation. Then, the appearance representation of foreground is superseded by that of background to accomplish the goal of image harmonization.

Inspired by traditional image harmonization methods \cite{xue2012understanding,multi-scale} which applied color transformation to adjust the foreground appearance, \citet{cong2022high} proposed to learn color transformation using deep learning for image harmonization. They combined color-to-color transformation and pixel-to-pixel transformation in a unified framework coherently. Several other works \cite{DCCF,liang2021spatial,RenECCV2022,Harmonizer,PCTNet,WangCVPR2023,menghigh} also proposed to predict various types of color transformations (\emph{e.g.}, color filter, rendering curve, linear transformation) for efficient image harmonization. Beyond different color transformations, some works~\cite{xu2023learning,DucoNet} explored different color spaces. 

Some works \cite{GiftNet,ShenICCV2023,HDNet} concurred that dynamic kernels applied to feature maps can boost the harmonization performance. Furthermore, \cite{GiftNet,ShenICCV2023} pointed out the importance of global information in dynamic kernel prediction. \citet{SycoNet} studied domain adaptive image harmonization by treating different datasets as different domains. Specifically,  an automatic augmentation network was developed to enrich the illumination diversity of a target domain with limited data.

Recently, some diffusion-based image harmonization models  \cite{li2023image,chen2023zero,zhou2024diffharmony,zhou2024diffharmony++,ren2024relightful,zhang2025scaling,liu2025dreamlight} have applied conditional diffusion model to image harmonization task. 
\citet{zhou2024diffharmony++} proposed to modify VAE decoder to alleviate the image distortion issue of diffusion model. \citet{ren2024relightful} proposed to inject background illumination information into diffusion model. \citet{zhang2025scaling} designed a diffusion model according to the  principle that the linear blending of an object’s appearances under different illumination conditions is consistent with its appearance under mixed illumination. \citet{chadebec2025lbm} proposed latent bridge matching between source image and target image, achieving superior and stable performance for image harmonization. 

\citet{tao2024diverse} pointed out that when transferring the background illumination to the foreground, different foreground reflectances would yield different harmonization results. They designed a reflectance-guided harmonization network,
which can produce diverse harmonized results considering different foreground reflectances. 

\subsection{Variants of Image Harmonization Task}

In this subsection, we discuss two variants of standard image harmonization task.

\textbf{Blind image harmonization: }Most image harmonization methods require the foreground mask as input, which means that the inharmonious region is known in advance. However, in real-world applications, we may not know the exact inharmonious region in advance. Image harmonization without foreground mask is called blind image harmonization. \citet{xiaodong2019improving} considered the problem of blind image harmonization. They proposed to predict the inharmonious region mask in the attention block, which deals with the foreground and background separately according to the predicted mask. 

Subsequently, some works \cite{liang2021inharmonious,liang2022inharmonious,AustNet,RSRNet,zhang2023multi,shin2025disharmony,chen2025novel} focused on inharmonious region localization task, which aims to localize the inharmonious region in an image. \citet{liang2021inharmonious} explored aggregating multi-scale contextual information and suppressing redundant information. The methods \cite{liang2022inharmonious,AustNet} proposed to magnify the domain discrepancy between foreground and background using color mapping for ease of identifying the inharmonious region. \citet{chen2025novel} explored multi-view representation including frequency view and flipped view. 

\begin{figure*}[t]
\begin{center}
\includegraphics[width=.92\linewidth]{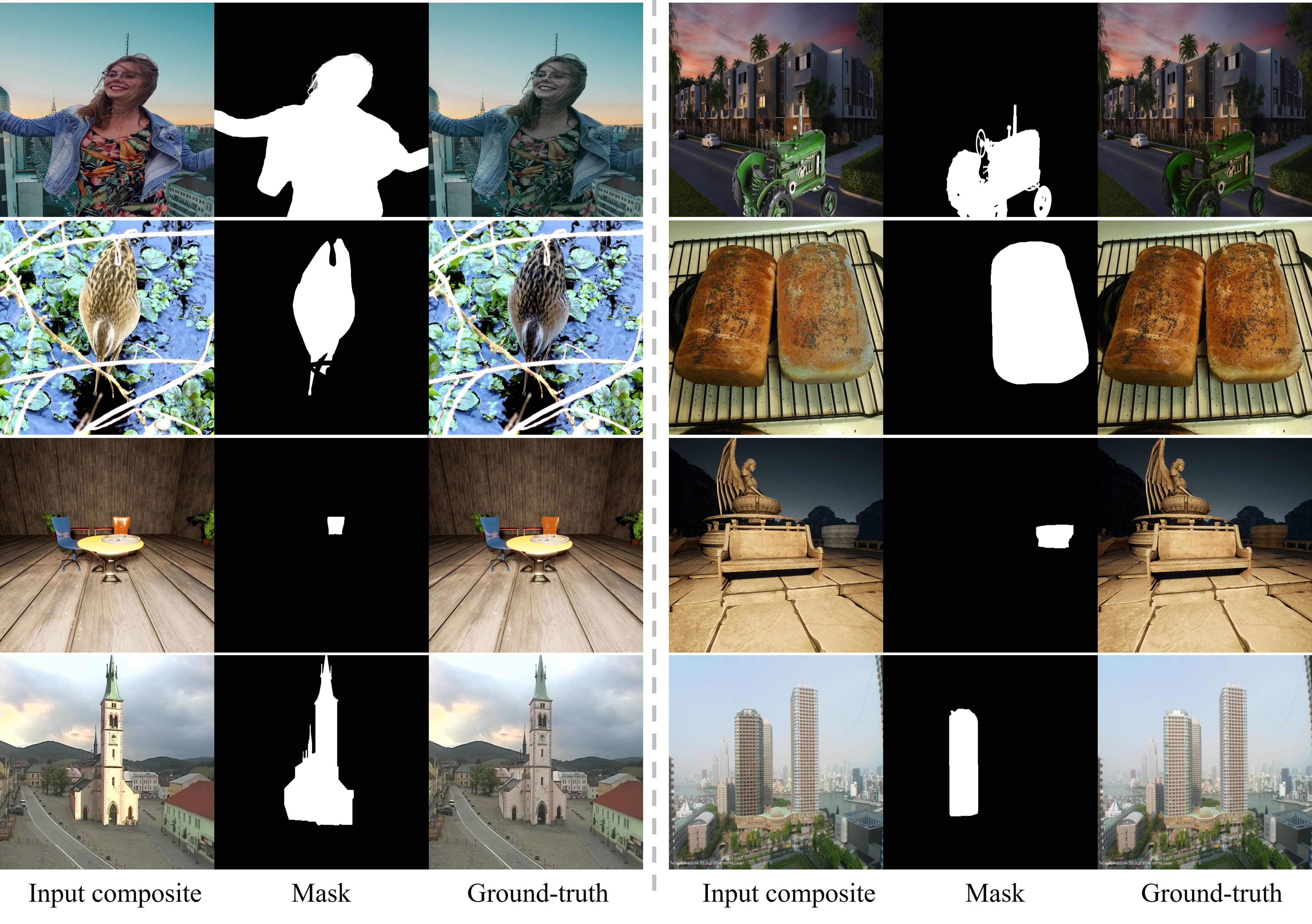}
\end{center}
\caption{In the first (\emph{resp.}, second, third, fourth) row, we show two examples from RealHM \cite{jiang2021ssh} (\emph{resp.}, HFlickr in iHarmony4 \cite{cong2021deep}, HVIDIT \cite{intriharm}, Hday2night in iHarmony4 \cite{cong2021deep}) dataset. From left to right in each example, we show the composite image, the foreground mask, and the ground-truth harmonized image. 
}
\label{fig:image_harmonization_examples}
\end{figure*}

\textbf{Painterly image harmonization: }In standard image harmonization, both foreground and background are from realistic images. There exist certain application scenarios that the background is an artistic image while the foreground is from a realistic image, in which case the standard image harmonization models may not work well. To overcome this problem, painterly image harmonization \cite{luan2018deep} has been studied to harmonize the realistic foreground according to the artistic background to obtain a uniformly stylized composite image. 

The relation between painterly image harmonization and standard image harmonization is like the relation between photorealistic style transfer and artistic style transfer. 
Painterly image harmonization is more challenging because multiple levels of styles (\emph{i.e.}, color, simple texture, complex texture)~\cite{niu2023progressive} need to be transferred from background to foreground, while  standard image harmonization only needs to transfer low-level style (\emph{i.e.}, illumination). 
Painterly image harmonization is also referred to as cross-domain image composition~\cite{HachnochiArXiv2023,TFICON,RefPaint,li2025aicomposer,wang2024primecomposer,pham2024tale}. 

The existing painterly image harmonization methods~\cite{luan2018deep,peng2019element,PHDNet,PHDiffusion,wang2024painterly,niu2023progressive,niu2023painterly} can be roughly categorized into optimization-based methods and feed-forward methods. 
Optimization-based methods optimize the input image to minimize the style loss and content loss, which is very time-consuming. 
For example, \citet{luan2018deep} proposed to optimize the input image with two passes, in which the first pass aims at robust coarse harmonization and the second pass targets at high-quality refinement. \citet{li2023freepih} proposed to optimize the latent features of diffusion model based on content loss and style loss.

Feed-forward methods send the input image through the model to output the harmonized result. For example, \citet{peng2019element} applied adaptive instance normalization to match the means and variances between the feature map of composite image and that of artistic background. \citet{PHDNet} performed painterly image harmonization in both frequency domain and spatial domain, considering that artistic paintings often have periodic textures and patterns which appear regularly.  \citet{niu2023progressive} divided styles into low-level styles (\emph{e.g.}, color, simple pattern) and high-level styles (\emph{e.g.}, complex pattern), and devised a progressive network which can harmonize a composite image from low-level styles to high-level styles progressively. \citet{niu2023painterly} proposed style-level supervision based on pairs of artistic objects and photographic objects, considering that it is hard to obtain pixel-wise supervision based on pairs of artistic objects and photographic objects. To achieve this goal, \citet{niu2023painterly} built an artistic object dataset containing the artistic objects segmented from artistic images. Each artistic object is associated with a list of photographic objects that have similar appearance and semantics to it. For each artistic object in an artistic image, it can be covered by a similar photographic object, yielding a composite image. Then, the harmonized photographic object in the composite image is expected to have the same style as the artistic object. 
\citet{sun2024painterly} applied dynamic kernel to painterly harmonization.
\citet{PHDiffusion} is the first work introducing diffusion model to painterly image harmonization, which can significantly outperform GAN-based methods when the background has dense textures or abstract style.

\subsection{Related Research Fields}
Image harmonization is closely related to style transfer. Note that both artistic style transfer \cite{gatys2015neural,huang2017arbitrary,park2019arbitrary} and photorealistic style transfer \cite{luan2017deep,li2018closed} belong to style transfer. Image harmonization is closer to photorealistic style transfer, which  transfers the style of a reference photo to another input photo. There are two main differences between image harmonization and photorealistic style transfer. 1) Firstly, image harmonization adjusts the foreground appearance according to the background, which must take the foreground location into consideration due to the locality property. In contrast, photorealistic style transfer adjusts the appearance of a whole input image according to another whole reference image. 2) Secondly, the definition of ``style" in photorealistic style transfer is unclear and coarsely depends on the employed style loss (\emph{e.g.}, Gram matrix loss \cite{gatys2015neural}, AdaIn loss \cite{huang2017arbitrary}). Differently, the goal of image harmonization is clearly adjusting the illumination statistics of foreground, so that the resultant foreground looks like the same object captured in the background illumination condition. 

\begin{figure*}[h]
\begin{center}
\includegraphics[width=.93\linewidth]{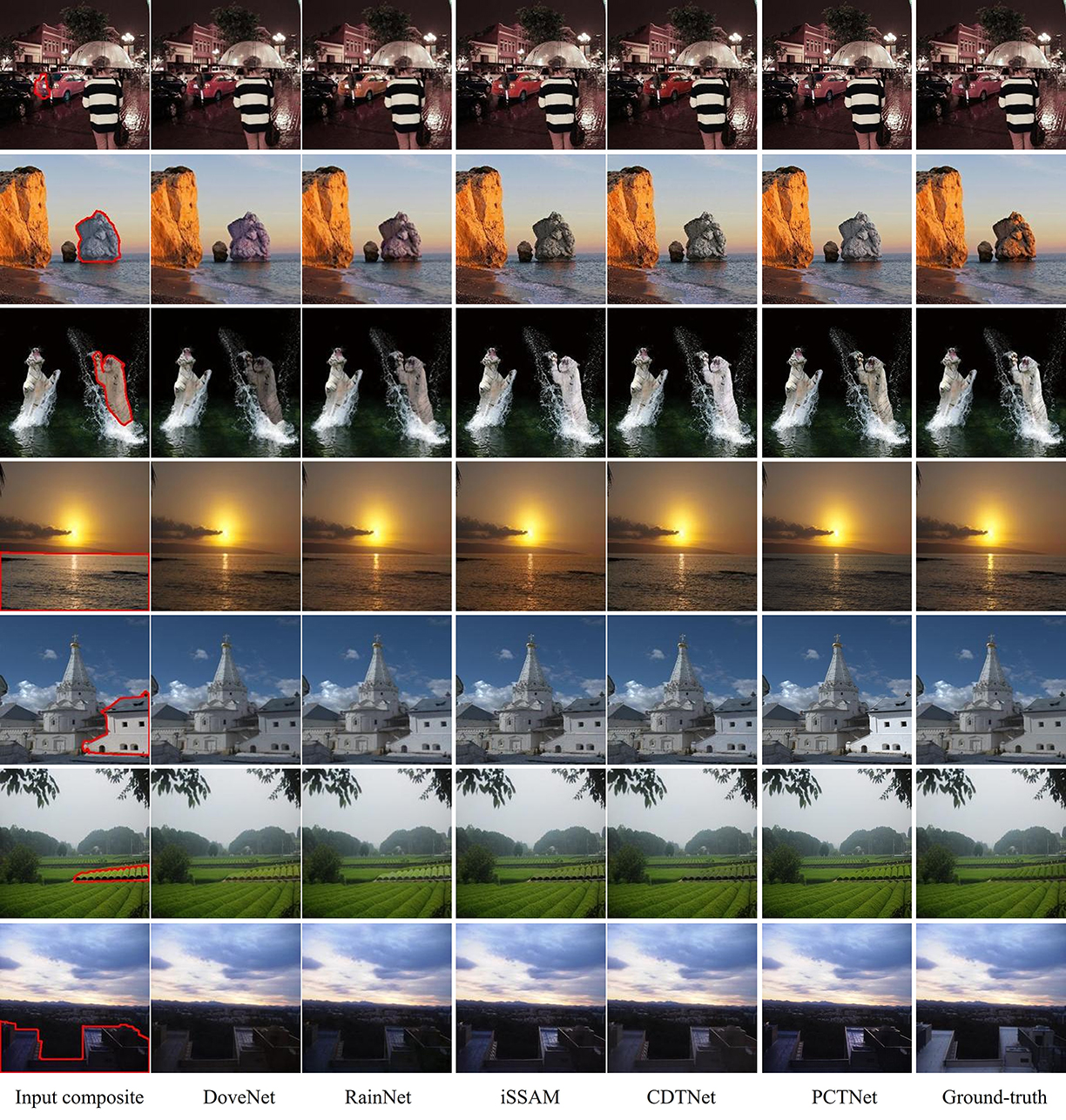}
\end{center}
\caption{The visualization results of different image harmonization methods on iHarmony4 \cite{DoveNet2020} dataset. From left to right in each row, we show the input composite image, the harmonization results of
DoveNet~\cite{DoveNet2020}, RainNet~\cite{regionaware},  iSSAM~\cite{sofiiuk2021foreground}, CDTNet~\cite{cong2022high}, PCTNet~\cite{PCTNet}, and the ground-truth harmonized image. }
\label{fig:image_harmonization_results}
\end{figure*}

\begin{figure*}[h]
\begin{center}
\includegraphics[width=.97\linewidth]{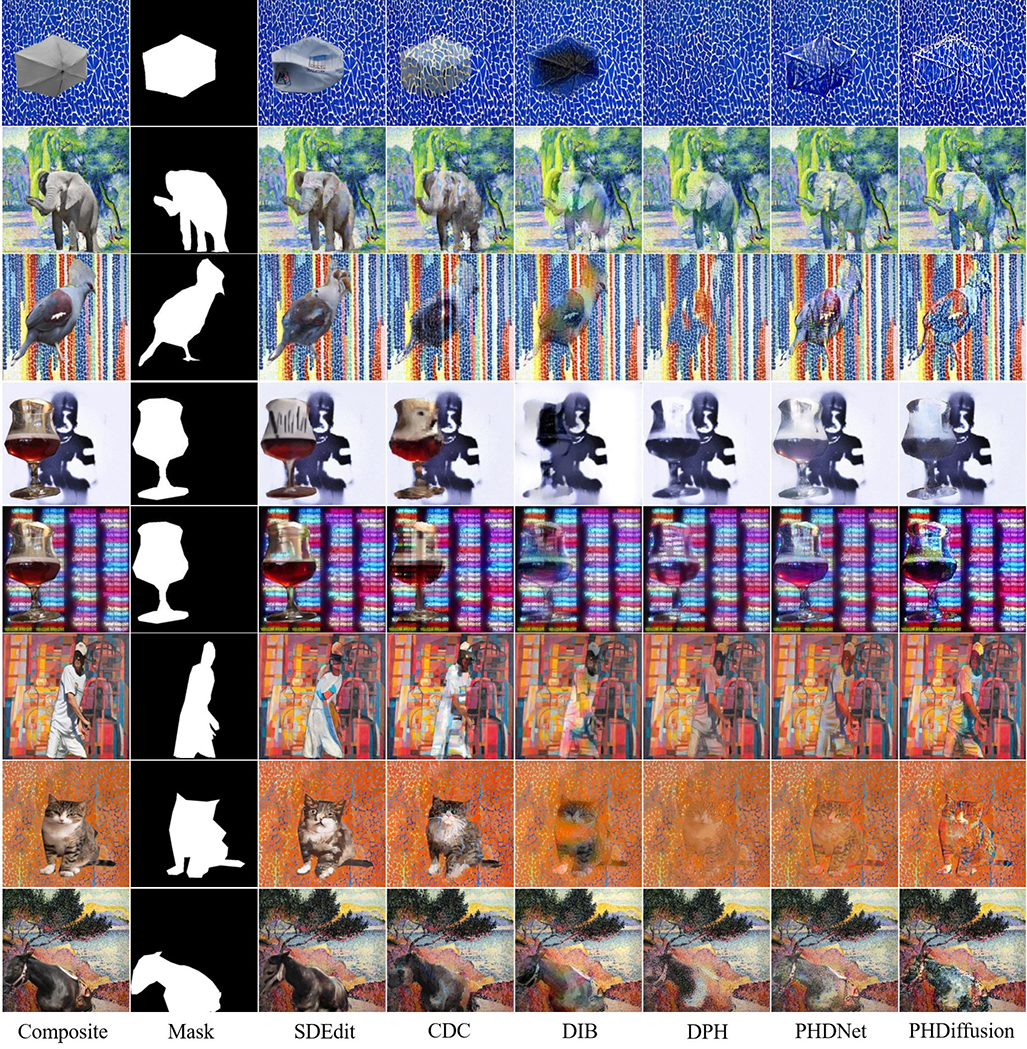}
\end{center}
\caption{The visualization results of different painterly image harmonization methods. From left to right in each row, we show the input composite image, the composite mask, the harmonization results of
SDEdit~\cite{meng2021sdedit}, CDC~\cite{HachnochiArXiv2023}, DIB~\cite{zhang2020deep}, DPH~\cite{luan2018deep}, PHDNet~\cite{PHDNet}, and PHDiffusion~\cite{PHDiffusion}. }
\label{fig:painterly_image_harmonization_results}
\end{figure*}

\subsection{Datasets and Evaluation Metrics} \label{sec:image_harmonization_dataset}

A large amount of composite images can be easily obtained by pasting the foreground from one image on another background image, but it is not easy to obtain the ground-truth harmonized image for the composite image. 
Training deep learning models requires abundant pairs of composite images and ground-truth harmonized images. Existing works have designed different schemes to construct image harmonization dataset. 
The key lies in how to construct a set of images with the same content yet different illuminations. 
We categorize the existing schemes into four groups: manual editing, color transfer, rendering technique, real shot. We show one representative dataset from each group in Fig.~\ref{fig:image_harmonization_examples}.

\textbf{Manual editing: }HAdobe5k sub-dataset in \cite{DoveNet2020} is constructed based on MIT-Adobe FiveK dataset \cite{bychkovsky2011learning}, in which each image is manually edited to five different illumination conditions. Swapping their foregrounds can yield pairs of composite images and ground-truth images. 
\citet{jiang2021ssh} released a small-scale \emph{RealHM} dataset with 216 image pairs, which is constructed based on real composite images obtained by pasting foregrounds on backgrounds. Human annotators manually adjust the foreground according to the background to obtain the ground-truth. Manual editing is time-consuming, labor-intensive, and unreliable. 

\textbf{Color transfer: }Some works \cite{tsai2017deep,xiaodong2019improving,DoveNet2020}  adjusted the foreground of real image to create synthetic composite image. Specifically, they treat a real image as harmonized image, segment a foreground region, and adjust this foreground region to be inconsistent with the background, yielding a synthetic composite image. HCOCO (\emph{resp.}, HFlickr) sub-dataset in \cite{DoveNet2020} is built upon COCO \cite{lin2014microsoft} (\emph{resp.}, crawled images from Flickr website), in which the foregrounds in real images are adjusted using traditional color transfer methods \cite{reinhard2001color,xiao2006color,fecker2008histogram,pitie2007automated}. These color transfer methods require reference object. Given a foreground object, an object from the same category and with similar appearance is found as its reference object. Then, color transfer is performed to change the illumination of foreground object to match the reference object. 
It is worth noting that such color transfer may produce low-quality synthetic composite images. Thus, \citet{DoveNet2020} manually filter out the low-quality synthetic composite images. SycoNet~\cite{SycoNet} learned a mapping from real images to filtered synthetic composite images, which can capture the human filtering knowledge and produce high-quality synthetic composite images. 

Another issue is that traditional color transfer methods may not faithfully reflect the natural illumination variation. To address this issue, \citet{GiftNet} proposed to transit across different illumination conditions by virtue of color checker, leading to ccHarmony dataset which can more faithfully reflect the natural illumination variation. Color checker is a simple way to record the illumination condition when taking photos. Given two photos with color checkers, color transfer parameters can be calculated based on two color checkers to transit across two illumination conditions. 

\textbf{Rendering technique: } \citet{cong2021deep} constructed \emph{RdHarmony} dataset by varying the lighting condition of the same scene using 3D rendering techniques. Within a set of images with the same scene yet various lighting conditions, the composite images could be obtained by exchanging the foregrounds between two images. Similarly, \citet{intriharm} constructed \emph{HVIDIT} dataset based on the rendered dataset \cite{elhelou2020vidit}. Another solution is rendering the same 3D foreground model using different illumination maps~\cite{pandey2021total,bao2022deep,hu2024sidnet,hu2024spatially}. However, the rendered images have a large domain gap with real images, so the harmonization model trained on rendered images cannot be directly applied to real test images. 

\textbf{Real shot: } A natural way to build image harmonization dataset is collecting a set of foreground images captured in different illumination conditions, followed by replacing one foreground with another counterpart. 
For example, Transient Attributes Database \cite{laffont2014transient} contains 101 sets, in which each set has well-aligned images for the same scene captured in different conditions (\emph{e.g.}, weather, time of the day, season). This dataset has been used to construct pairs in Hday2night sub-dataset in \cite{DoveNet2020}. However, collecting the dataset like  \cite{laffont2014transient} calls for capturing the same scene with a fixed camera for a long time, which is hard to be realized in practice. To obtain the foregrounds in different capture conditions,  \citet{song2020illumination} proposed an interesting way to construct \emph{GMS} Dataset. Specifically, they place the same physical model (3D foreground object) in different lighting conditions to capture different images and align the foregrounds in different images. Nevertheless, the collection cost is still very high and the diversity of foreground is restricted. For indoor scenes, another solution is using light stage (a large number of individual lights placed around the scene) to manually control the illumination~\cite{murmann2019dataset}, so that one scene can be quickly switched to different illumination conditions. 

Existing works adopt metrics including Mean Square  Error (MSE), Peak Signal-to-Noise Ratio (PSNR), Structural  SIMilarity  index (SSIM)~\cite{schieber2017quantification}, Learned Perceptual Image Patch Similarity (LPIPS)~\cite{zhang2018unreasonable} to calculate the distance between harmonized result and ground-truth. These metrics can also be calculated only within the foreground region. Besides, they conduct user study on real composite images by asking engaged users to select the most realistic images and calculate the metric (\emph{e.g.},  B-T score \cite{bradley1952rank}, ratio).

\subsection{Experiments}

We conduct experiments for both standard image harmonization and painterly image harmonization.

For standard image harmonization, we use iHarmony4 \cite{DoveNet2020} dataset (HCOCO, HFlickr, HAdobe5k, and Hday2night), which is the most commonly used dataset for image harmonization. All methods are trained on the combination of training sets from four sub-datasets, and evaluated on the test set from each sub-dataset. In Fig. \ref{fig:image_harmonization_results}, we show the harmonized results of different methods (DoveNet~\cite{DoveNet2020},    RainNet~\cite{regionaware},  iSSAM~\cite{sofiiuk2021foreground}, CDTNet~\cite{cong2022high}, PCTNet~\cite{PCTNet}). We observe that some competitive methods can generally produce visually appealing results that are close to the ground-truth images. However, when the background illumination is very complex or the composite foreground and background have dramatically divergent illumination statistics, the existing methods are still struggling to harmonize the foreground to approach the ground-truth. 

For practical usage, in usual lighting conditions,  color-to-color transformation methods like \cite{PCTNet} are recommended, because they can process high-resolution images efficiently and preserve the image details well. In unusual lighting conditions (\emph{e.g.}, neon light, high-contrast shadow caused by non-uniform lighting), diffusion-based methods may perform better. It is worth noting that among the diffusion-based methods, the methods (\emph{e.g.}, \cite{chadebec2025lbm}) adopting latent bridge matching between source image and target image generate remarkably stable results. 

For painterly image harmonization, we use COCO \cite{lin2014microsoft} and WikiArt \cite{nichol2016painter}.
COCO \cite{lin2014microsoft} contains instance segmentation annotations for 80 object categories, while WikiArt \cite{nichol2016painter} contains digital artistic paintings from different styles.
We create composite images based on these two datasets, with the photographic objects from COCO and the painterly backgrounds from WikiArt. In Fig. \ref{fig:painterly_image_harmonization_results}, we show the harmonized results of different methods (SDEdit~\cite{meng2021sdedit}, CDC~\cite{HachnochiArXiv2023}, DIB~\cite{zhang2020deep}, DPH~\cite{luan2018deep}, PHDNet~\cite{PHDNet}, PHDiffusion~\cite{PHDiffusion}). We split COCO and WikiArt into training set and test set, based on which all methods are trained and evaluated. 
It can be seen that the methods (DPH, PHDNet, PHDiffusion) specifically designed for painterly image harmonization significantly outperform the other methods. In the challenging cases, where the background has dense textures or abstract styles, PHDiffusion achieves remarkable performance, probably due to the generative ability of diffusion model and the rich prior knowledge in foundation model. 

For practical usage, in most cases, feed-forward GAN-based methods like \cite{PHDNet} can achieve satisfactory results. In the challenging cases such as dense textures or abstract style,  feed-forward diffusion-based methods like \cite{PHDiffusion} are able to demonstrate their strengths.

%% file: sections/shadow_generation.tex
\section{Shadow Generation}\label{sec:shadow_generation}

\begin{figure*}[t]
\begin{center}
\includegraphics[width=.97\linewidth]{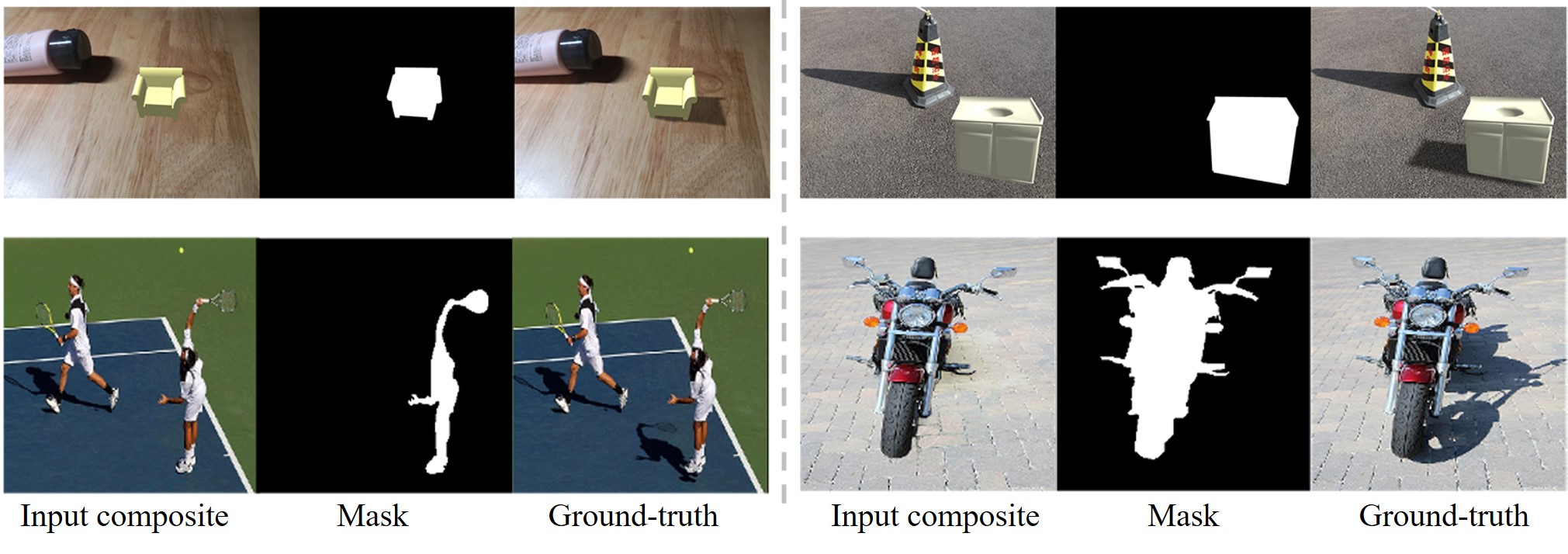}
\end{center}
\caption{In the first row, we show two examples from  Shadow-AR \cite{liu2020arshadowgan} dataset, which is constructed based on rendered images. In the second row, we show two examples from DESOBA \cite{hong2021shadow} dataset , which is constructed based on real images. From left to right in each example, we show the composite image without foreground shadow, the foreground mask, and the ground-truth image with foreground shadow. 
}
\label{fig:shadow_generation}
\end{figure*}

In the previous section, image harmonization methods could adjust the foreground appearance to make it compatible with the background, but they ignore the fact that the inserted object may also have impact on the background. For example, if background objects cast shadows on the ground but the inserted object does not have shadow, the composite image would look unrealistic. 
To address this issue, shadow generation task aims to generate plausible shadow for the foreground object according to background illumination information to make the composite image more realistic. Similar to Section~\ref{sec:image_harmonization}, we divide the existing methods into rendering based methods and non-rendering based methods. 

\subsection{Rendering based Methods}

The traditional methods \cite{karsch2011rendering,karsch2014automatic,liu2017static,liao2019illumination} usually use rendering techniques to generate shadow for the inserted foreground object, which need to collect or estimate the scene geometry, foreground object geometry, and scene illumination. For example, \cite{karsch2011rendering} proposed to collect the rough geometry information and lighting information from users, based on which rendering techniques could be employed. 
However, it is very tedious and sometimes impossible to collect all the required information.  
In \cite{karsch2014automatic,liu2017static,liao2019illumination}, they attempted to estimate the missing information (\emph{e.g.}, scene geometry, lighting information) automatically. With the recovered information, the local region to place the inserted 3D object is rendered with and without the inserted foreground object. The difference between these two rendered images reveals the impact of foreground object on the background, which is added to the input composite image to produce the target image with foreground shadow. Geometry estimation and lighting estimation based on a single image have been long studied, and many different technical approaches have been developed \cite{liu2017static,kronander2015photorealistic}. More recently, some methods~\cite{liao2019illumination,gardner2019deep,zhang2019all,garon2019fast,hold2019deep,weber2018learning} endeavored to estimate illumination condition and scene geometry based on a single image using deep learning models, which could achieve better performance than traditional estimation models. Some works~\cite{controlshadow,PixHtLab} proposed to forecast essential geometry information (\emph{e.g.}, pixel height) which cooperates with user-specified illumination to render realistic shadows. 

Despite the remarkable progress they have achieved, it is still very challenging to accurately estimate the geometry and lighting information in complex real-world scenes. Erroneous estimation may mislead the rendering process and produce terrible results~\cite{zhang2019shadowgan}.

\begin{figure*}
\begin{center}
\includegraphics[width=.99\linewidth]{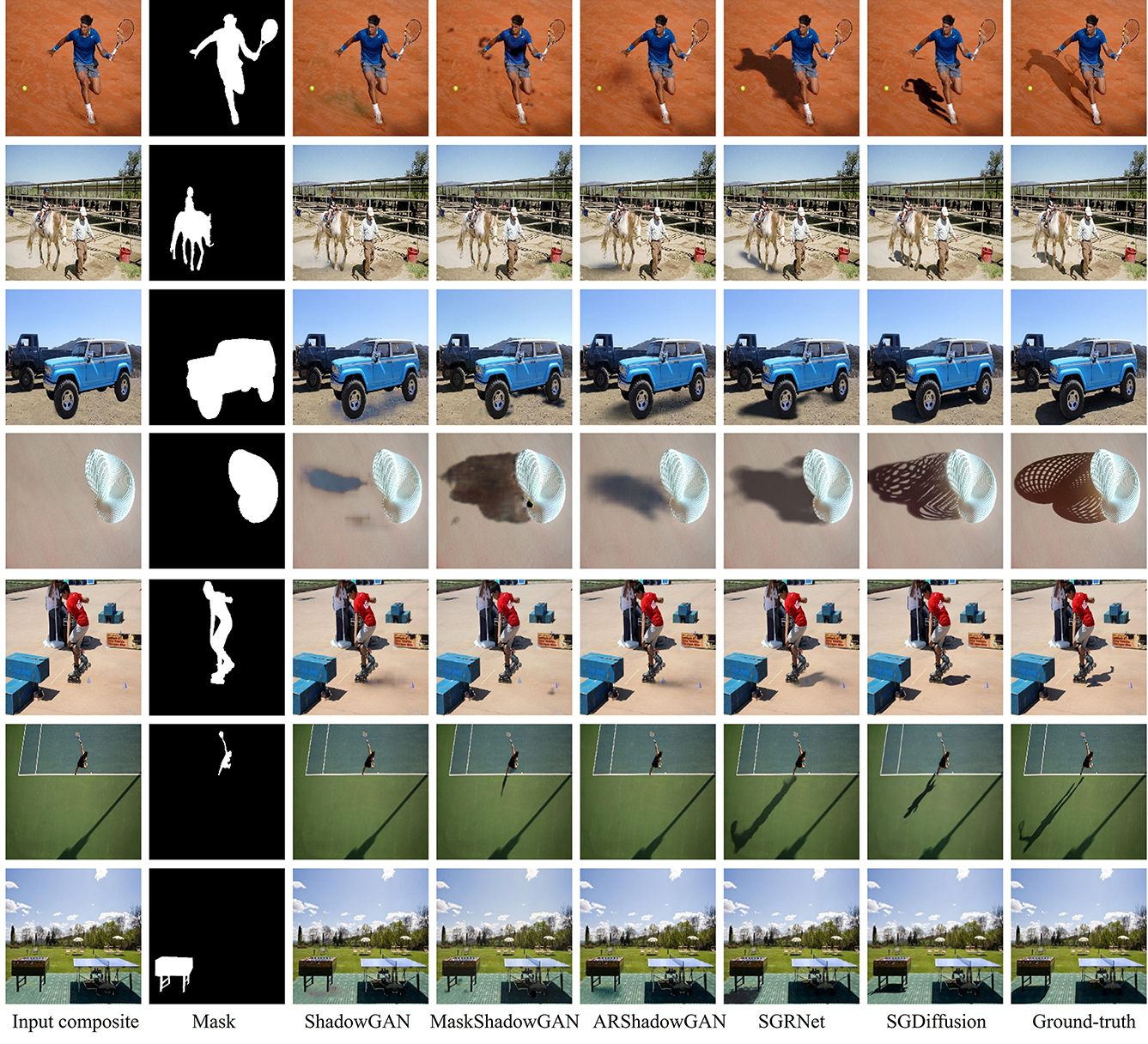}
\end{center}
\caption{The visualization results of different shadow generation methods on DESOBA \cite{hong2021shadow} dataset. From left to right in each row, we show the input composite image, the composite foreground mask,  the generated results of ShadowGAN~\cite{zhang2019shadowgan}, MaskShadowGAN~\cite{hu2019mask}, ARShadowGAN~\cite{liu2020arshadowgan}, SGRNet~\cite{hong2021shadow}, SGDiffusion~\cite{DESOBAv2}, and the ground-truth shadow image. }
\label{fig:shadow_generation_results} 
\end{figure*}

\subsection{Non-rendering based Methods}

Recently, some works treat shadow generation as an image-to-image translation task, and develop deep  networks which translate input composite image without foreground shadow to the target image with foreground shadow. For instance, \citet{zhan2020adversarial} used an auto-encoder to predict the shadow mask with a pretrained illumination model~\cite{gardner2017learning, illuminationdata} to provide illumination information. The generated images are pushed towards real images with foreground shadows using adversarial learning. 

Other methods \cite{zhang2019shadowgan,inoue2020rgb2ao,liu2020arshadowgan,hong2021shadow,fu20263d} utilized paired training data (paired images with and without foreground shadow) to generate better shadow images. ShadowGAN~\cite{zhang2019shadowgan} employed standard conditional GAN with reconstruction loss, local adversarial loss, and global adversarial loss to generate shadow for the inserted 3D foreground objects. 
\citet{inoue2020rgb2ao} developed a multi-task framework with two decoders accounting for depth map prediction and ambient occlusion map prediction respectively. 
ARShadowGAN~\cite{liu2020arshadowgan} proposed an attention-guided residual network. The network predicts two attention maps for background shadow and occluder respectively, which are concatenated with composite image and foreground object mask to produce a residual shadow image. 
SGRNet \cite{hong2021shadow} designed a two-stage shadow generation network. In the first stage, foreground features and background features are interacted using cross-attention to predict a shadow mask. In the second stage, they predict shadow parameters which are used to darken the input composite image. Then, the darkened image is combined with the input composite image with shadow matte. \citet{MengTMM2023} adopted a similar two-stage pipeline and proposed to generate the shadow region by fusing multiple underexposure images. DMASNet~\cite{RdSOBA} decomposed shadow mask prediction into  box prediction and shape prediction, followed by attending relevant background shadow pixels to fill in the predicted shadow region. The method in \cite{fu20263d} proposed a 3D-aware shadow generation model by embedding tri-plane feature representations into a pixel-aligned volume rendering pipeline. 
Some other shadow generation methods are not designed for our task, \emph{i.e.}, generating shadow for the foreground object in a composite image, but they can be somehow adapted to our task. Mask-ShadowGAN~\cite{hu2019mask} explored conducting shadow removal and shadow generation with unpaired data at the same time, which satisfies cyclic consistency. The shadow generation branch can be directly extended to generate foreground shadow. \citet{sheng2021ssn} designed a shadow generation network to generate soft shadow for foreground object with user control. They first predict ambient occlusion map, which is jointly used with user-provided light map to produce soft shadow mask. When adapted to our task, an environment light map needs to be inferred from background before using their network. 

SGDiffusion~\cite{DESOBAv2} is the first work on shadow generation using diffusion model, which is built upon ControlNet~\cite{zhang2023adding} with extra intensity module to refine the shadow intensity. \cite{winter2024objectdrop,tarres2024thinking,wang2025metashadow,ahmed2026coshadow} also trained conditional diffusion model for shadow generation. \cite{zhou2024foreground,yu2024cfdiffusion} first predicted coarse shadow mask and then fed the shadow mask to diffusion model. \citet{zhao2025shadow} injected geometry prior, \emph{i.e.}, shadow location and shape, into diffusion model to enhance the quality of generated shadows with complex shapes. 
The diffusion-based methods can generate reasonable shadows for foreground objects in the composite images with simple scene and illumination condition, but often fail to generate reasonable shadows for the composite images with complex scene and illumination condition. Moreover, the generated shadows have roughly correct locations and shapes, but lack realistic contours and details matching the foreground objects.

\subsection{Datasets and Evaluation Metrics}
Similar to image harmonization in Section~\ref{sec:image_harmonization}, composite images without foreground shadows can be easily obtained. Nonetheless, it is very difficult to obtain paired data, \emph{i.e.}, a composite image without foreground shadow and a ground-truth image with foreground shadow, which are required by supervised deep learning methods on shadow generation~\cite{zhang2019shadowgan,liu2020arshadowgan,hong2021shadow}. Some works~\cite{zhang2019shadowgan,liu2020arshadowgan} construct rendered datasets with paired data by inserting a virtual object into 3D scene and generating shadow for this object with rendering technique. 
ARShadowGAN~\cite{liu2020arshadowgan} released a rendered dataset named \emph{Shadow-AR} by inserting a foreground object into real background image and generating its corresponding shadow with rendering technique. Shadow-AR dataset contains $3,000$ quintuples, in which each quintuple consists of a composite image without foreground shadow,  its corresponding ground-truth image with foreground shadow, foreground object mask, background object mask, and background shadow mask. Shadow-AR dataset only uses 13 foreground objects from  ShapeNet~\cite{ShapenetChang2015}  and Stanford 3D scanning repository, so the diversity of dataset is very limited. 
Some examples in Shadow-AR dataset are exhibited in the first row in Fig.~\ref{fig:shadow_generation}, in which we show the composite image without foreground shadow, foreground object mask, and ground-truth image with foreground shadow. Similar to ARShadowGAN~\cite{liu2020arshadowgan},  ShadowGAN~\cite{zhang2019shadowgan} also adopted rendering technique to construct a rendered dataset, which uses $9,265$ foreground objects from  ShapeNet~\cite{ShapenetChang2015} and $110$ background textures (\emph{e.g.}, woolen, stone, tablecloth) collected from Internet. \citet{RdSOBA} contributed a large-scale rendering dataset called RdSOBA, which has 788 3D foreground objects and nearly 280,000 object-shadow pairs.
In particular, they place a group of 3D objects in
the 3D scene, and get the images without or with object
shadows using rendering techniques. 

Although it is feasible to generate paired data using rendering technique, the rendered images have large domain gap with real images. When applying the model trained on rendered images to real images, the performances are usually significantly degraded. To overcome this drawback, \citet{hong2021shadow} constructed paired data by manually removing the foreground shadows from real shadow images in SOBA dataset~\cite{wang2020instance} to produce synthetic composite images without foreground shadows, leading to \emph{DESOBA} dataset. This strategy to create synthetic composite images is similar to the backward adjustment for constructing image harmonization dataset (see Section~\ref{sec:image_harmonization}). In particular, \citet{hong2021shadow} first remove all shadows from a shadow image to create a shadow-free image. Then, one foreground shadow region in the shadow image is overlaid by the counterpart in its corresponding shadow-free image, yielding a synthetic composite image with one missing foreground shadow. DESOBA dataset contains 839 training images with totally 2,995 object-shadow pairs and 160 test images with totally 624 object-shadow pairs.
Some examples in DESOBA dataset are exhibited in the second row in Fig.~\ref{fig:shadow_generation}, in which we show the composite image without foreground shadow, foreground object mask, and ground-truth image with foreground shadow. As mentioned in \cite{hong2021shadow}, manual shadow removal is extremely expensive. 

To alleviate the burden of manually annotating masks and removing shadows, \cite{DESOBAv2} design an automatic pipeline to construct shadow generation dataset and contributed a larger-scale dataset DESOBAv2. Specifically, \cite{DESOBAv2} employ the pretrained object-shadow detection model \cite{detect4} to predict object-shadow masks and employ the off-the-shelf inpainting model \cite{rombach2022high} to inpaint the shadow regions. DESOBAv2 has 21,575 images with 28,573 valid object-shadow pairs. 

Instead of constructing synthetic datasets \cite{hong2021shadow,DESOBAv2} by removing the shadows, 
\cite{winter2024objectdrop,kim2025orida} constructed real-world dataset by taking photos with object (factual image) or without object (counterfactual image). However, this approach to construct dataset is very costly and labor-intensive.

To evaluate the quality of generated composite images with foreground shadows, 1) existing shadow generation works \cite{zhan2020adversarial} without paired data adopt Frechet Inception Distance (FID) \cite{FIDHeusel2017} and Manipulation Score (MS)~\cite{chen2019toward} to measure the realism of generated shadow images. 2) For the works \cite{liu2020arshadowgan,hong2021shadow} with paired data, they adopt Structural SIMilarity index (SSIM)~\cite{schieber2017quantification} and Root Mean Square Error (RMSE)~\cite{barron2014shape} to measure the difference between generated image and ground-truth image. SSIM and RMSE can also be calculated only within the ground-truth foreground shadow region. 
\citet{liu2020arshadowgan} also use Balanced Error Rate (BER)~\cite{nguyen2017shadow} to evaluate the quality of predicted shadow mask based on ground-truth shadow mask. 3) User study is also called for to ensure that generated shadows comply with human perception.  Participants are asked to select the most realistic images, based on which some metrics (\emph{e.g.},  B-T score \cite{bradley1952rank}, ratio) are calculated. 

\subsection{Experiments}

We compare existing shadow generation methods ShadowGAN~\cite{zhang2019shadowgan}, MaskShadowGAN~\cite{hu2019mask}, ARShadowGAN~\cite{liu2020arshadowgan}, SGRNet~\cite{hong2021shadow}, and SGDiffusion~\cite{DESOBAv2}. All methods are trained on the training set of DESOBA~\cite{hong2021shadow} and DESOBAv2~\cite{DESOBAv2}, and evaluated on the test set of DESOBA. 
We show the shadow images generated by different methods in Fig.~\ref{fig:shadow_generation_results}. 

It can be seen that most methods \cite{zhang2019shadowgan,hu2019mask,liu2020arshadowgan} are struggling to produce reasonable shadow for the foreground object, or even produce no shadow at all, which implies that shadow generation for the inserted foreground object is a very tough task. SGRNet~\cite{hong2021shadow} achieves relatively compelling results, but the shapes of generated shadows are often unrealistic. Besides, we observe that SGRNet tends to overfit the artifacts caused by manual shadow removal in DESOBA training set, leading to the results perfectly matching the ground-truth (\emph{e.g.}, row 6, 7).
SGDiffusion~\cite{DESOBAv2} obtains the most competitive results by resorting to the foundation diffusion model, even for the foreground objects (\emph{e.g.}, row 1, 4) with complicated shapes, and demonstrates remarkable generalization ability. 

For practical usage, foundation diffusion model (\emph{e.g.}, SD~\cite{rombach2022high}, FLUX~\cite{flux}) is imperative to generate realistic shadows. More advanced foundation models usually lead to better performance, especially in the challenging cases.

%% file: sections/reflection_generation.tex
\section{Reflection Generation}\label{sec:reflection_generation}

\begin{figure*}[t]
\begin{center}
\includegraphics[width=.9\linewidth]{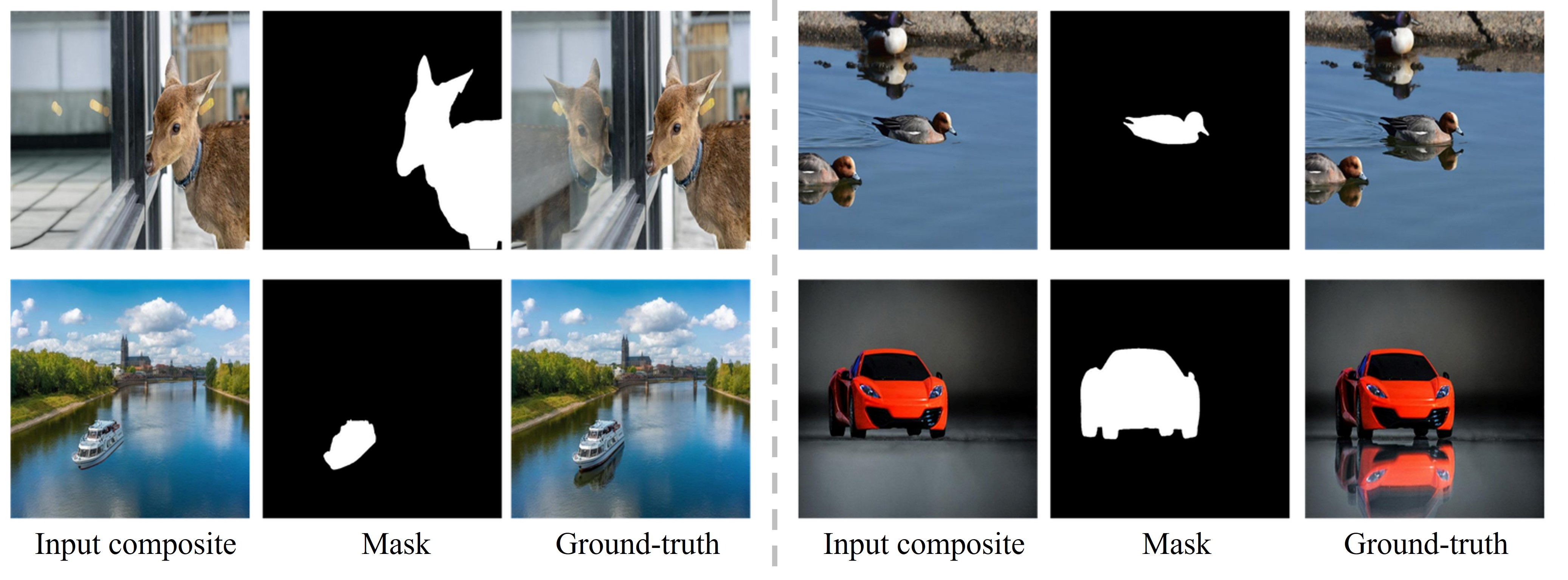}
\end{center}
\caption{We show four examples from DEROBA~\cite{rgdiffusion} dataset. From left to right in each example, we show the composite image without foreground reflection, the foreground mask, and the ground-truth image with foreground reflection. 
}
\label{fig:reflection_generation}
\end{figure*}

Besides shadow, reflection is another impact that the foreground object may cast on the background, especially when the foreground object is above the water or on the desktop with reflective material.
To address this issue, reflection generation task aims to generate plausible reflection for the foreground object to make the composite image more realistic. Similar to Section~\ref{sec:shadow_generation}, we divide the existing methods into rendering based methods and non-rendering based methods. 
       '
\subsection{Rendering based Methods}

\citet{ma2021neural} first estimated the lighting information and then employed a renderer to generate coarse reflection, which is further refined using neural network. 

\subsection{Non-rendering based Methods}

Some works treat reflection generation as image-to-image translation task without estimating lighting/geometry information or employing renderer.  For example, the works \cite{winter2024objectdrop,tarres2024thinking} employed conditional diffusion model to synthesize reflection for the inserted foreground object. The works \cite{dhiman2024reflecting,dhiman2025mirrorverse} adopted similar approaches but focused on mirror reflection. RGDiffusion~\cite{rgdiffusion} extended ControlNet~\cite{zhang2023adding} considering the property of reflection generation task. Specifically, they inject the cropped foreground into denoising UNet via cross-attention, considering that reflections are usually the horizontal or vertical mirror images of foreground objects.

\subsection{Datasets and Evaluation Metrics}

Training conditional diffusion model for reflection synthesis requires paired training data, \emph{i.e.}, a composite image without foreground reflection and a ground-truth image with foreground reflection. 

Some works~\cite{dhiman2024reflecting,dhiman2025mirrorverse} constructed rendered datasets with paired data by inserting a virtual object into 3D scene and generating reflection in the mirror with rendering technique. For example, \citet{dhiman2024reflecting} released SynMirror dataset with 66,068 objects and 198,204 rendered images. 

Different from rendering pipeline and similar to the pipeline of constructing shadow generation dataset DESOBAv2 \cite{DESOBAv2}, DEROBA~\cite{rgdiffusion} manually annotated object-reflection masks and employed the off-the-shelf inpainting model to inpaint the reflection regions, resulting in synthesized composite images. Some
examples in DEROBA dataset are exhibited in Fig.~\ref{fig:reflection_generation}, in which we show the composite image without foreground reflection, foreground object mask, and ground-truth
image with foreground reflection.

Instead of constructing synthetic datasets \cite{dhiman2024reflecting,dhiman2025mirrorverse}, some other works
\cite{winter2024objectdrop,kim2025orida} constructed real-world dataset by taking photos with object (factual image) or without object (counterfactual image). However, this approach to construct dataset is very costly and labor-intensive. 
 
To evaluate the quality of generated composite images with foreground reflections, Mean Squared  Error (MSE), Peak Signal-to-Noise Ratio (PSNR), Structural SIMilarity index (SSIM)~\cite{schieber2017quantification}, Learned Perceptual Image Patch Similarity (LPIPS)~\cite{zhang2018unreasonable} can be adopted to measure the difference between generated image and ground-truth image.  User study is also necessary to ensure that generated reflections conform to human perception. Participants are asked to select the most realistic images, based on which some metrics (\emph{e.g.},  B-T score \cite{bradley1952rank}, ratio) are calculated.

\subsection{Experiments}

\begin{figure*}[t]
\begin{center}
\includegraphics[width=.98\linewidth]{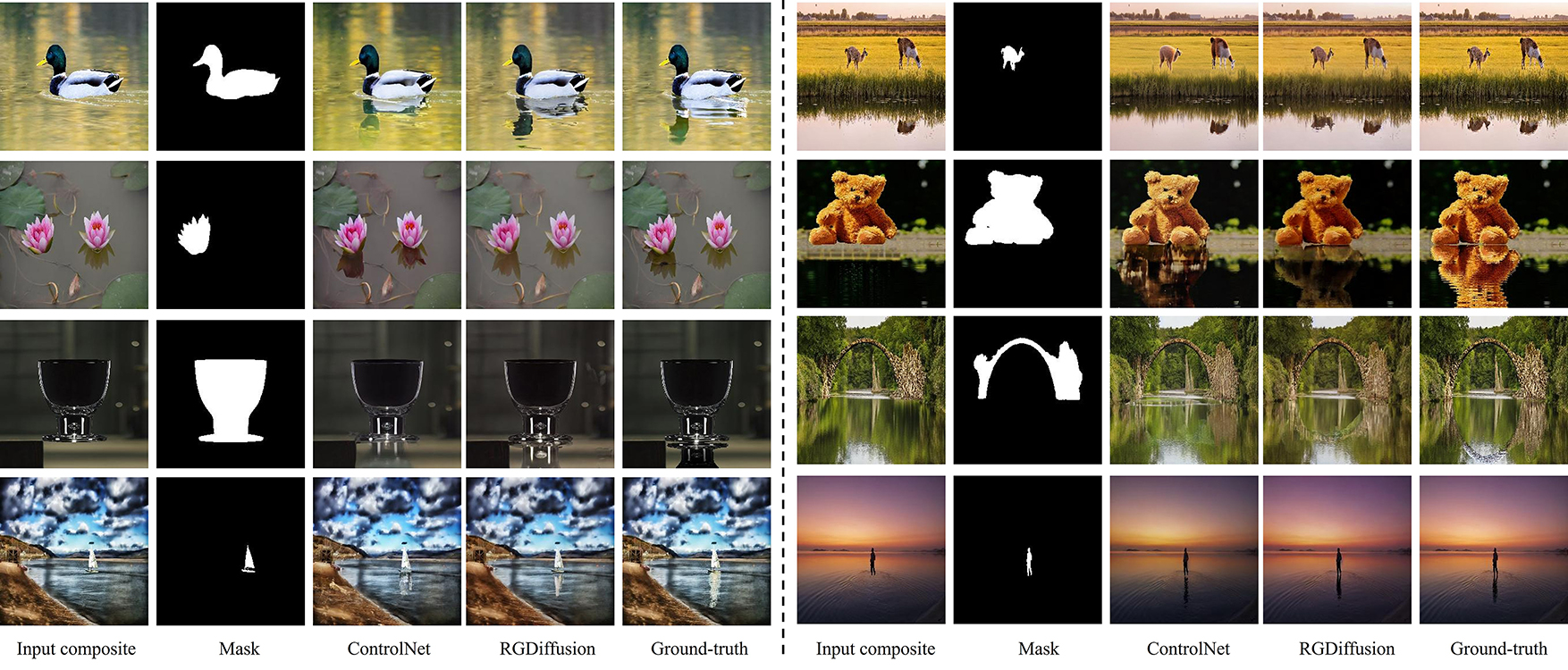}
\end{center}
\caption{The visualization results of different reflection generation methods on DEROBA~\cite{rgdiffusion} dataset. From left to right in each group, we show the input composite image, the composite foreground mask,  the generated results of ControlNet~\cite{zhang2023adding}, RGDiffusion~\cite{rgdiffusion}, and the ground-truth reflection image. }
\label{fig:reflection_generation_exp}
\end{figure*}

We compare different methods~\cite{zhang2023adding,rgdiffusion} on DEROBA~\cite{rgdiffusion} dataset. As shown in Fig.~\ref{fig:reflection_generation_exp},  RGDiffusion~\cite{rgdiffusion} produces more accurate shapes and clearer details compared with ControlNet~\cite{zhang2023adding}.  

For practical usage, the methods built upon foundation diffusion model (\emph{e.g.}, SD~\cite{rombach2022high}, FLUX~\cite{flux}) can usually produce satisfactory results. 

%% file: sections/generative_composition.tex
\section{Generative Composition}\label{sec:generative_composition}

 \begin{figure*}[t]
    \begin{center}
    \includegraphics[width=.99\linewidth]{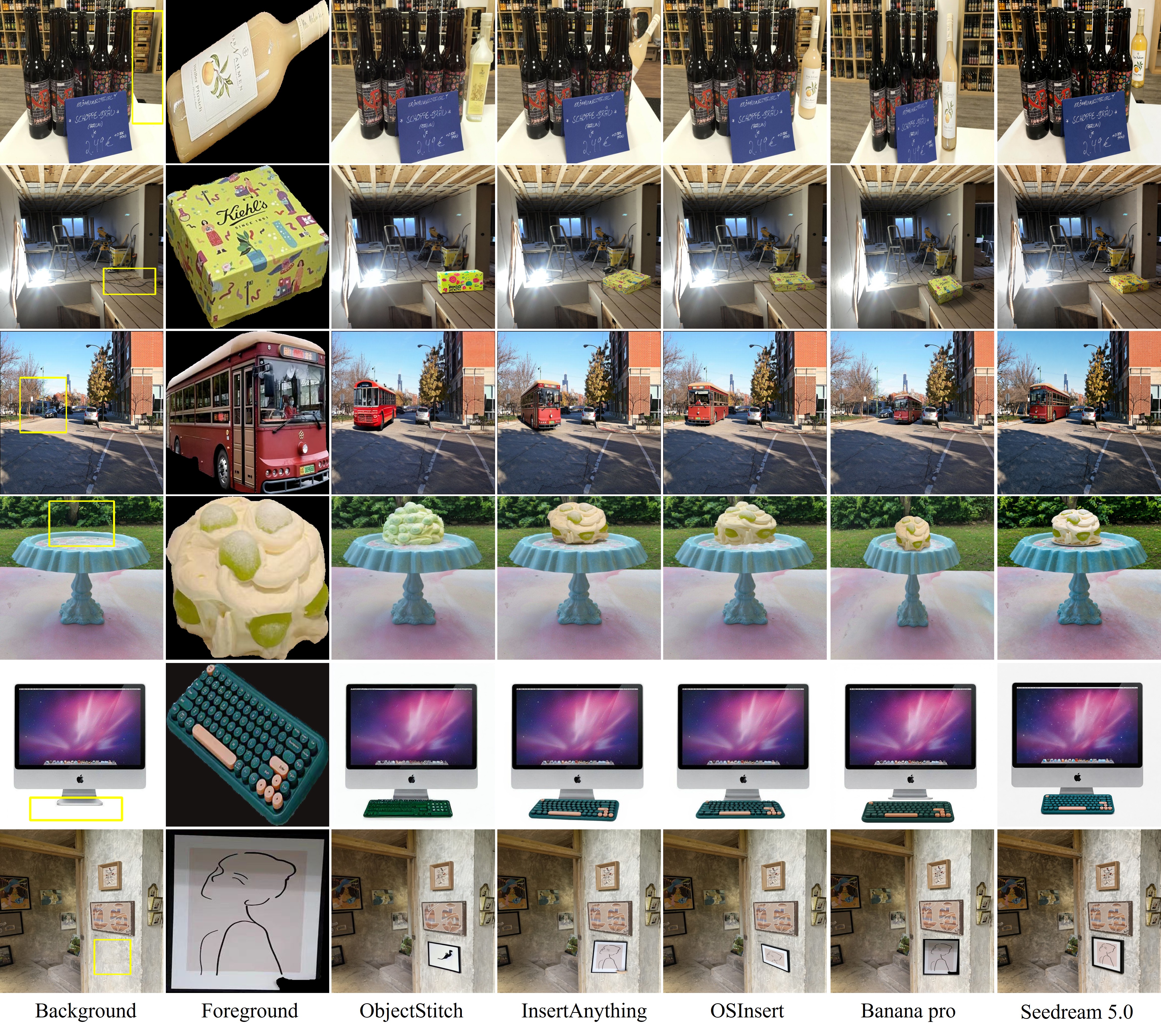}
    \end{center}
    \caption{The visualization results of different generative composition methods on MureCom~\cite{lu2023dreamcom} dataset.
    From left to right in each row, we show the background with foreground bounding box, five reference images of the same foreground object,  the generated results of ObjectStitch~\cite{objectstitch}, InsertAnything~\cite{song2025insert}, OSInsert~\cite{osinsert}, Banana pro~\cite{team2024gemini}, and Seedream 5.0.}
    \label{fig:generative_composition_results}
\end{figure*}

As the diffusion models \cite{rombach2022high} pretrained on large-scale dataset~\cite{Laion400m} become popular in various image generation and editing tasks, generative image composition (object compositing) has attracted growing research interest. 
In contrast with previous methods which perform one or multiple sub-tasks sequentially, generative image composition is a combinatorial task which performs multiple sub-tasks (\emph{e.g.}, image blending, image harmonization, shadow generation) in parallel through one unified model. Given a foreground, a background, and a bounding box indicating the foreground placement, generative image composition aims to directly produce a realistic composite image with the foreground naturally and harmoniously merged into the background. 

Generative image composition has certain overlap with object-guided image inpainting \cite{PBE} and image customization \cite{yuan2023customnet}. Their differences are claimed as follows. 1) Object-guided image inpainting needs a mask to indicate the inpainted region, where the mask shape usually implies the target shape of inserted object. When the inpainted region is a bounding box free of shape information, object-guided image inpainting is closer to generative image composition. However, strictly speaking, generative image composition expects to preserve the non-foreground pixels in the bounding box, which is different from object-guided image inpainting. Moreover, generative image composition aims to generate shadow and reflection for the foreground object without box or shape constraint, which is also different from object-guided image inpainting.
2) Image customization is a very broad concept, which includes changing attributes and adding background for a specific object. Generative image composition can be deemed as a special case of image customization. 

\subsection{Deep Learning Methods}
The existing generative image composition methods can be divided into two groups: training-free methods and training-based methods.

\subsubsection{Training-free Methods}

The first group of methods \cite{HachnochiArXiv2023,TFICON,wang2024primecomposer,li2024tuning,xu2025context,lu2025does} utilizes off-the-shelf foundation generation model, which does not require training or finetuning. They aim to generate high-quality composite images by manipulating the foreground and background elements (\emph{e.g.}, feature, attention) through the denoising process.

\subsubsection{Training-based Methods}

The second group of methods \cite{PBE,objectstitch,PhDZhang,zhang2023controlcom,carecom,yuan2023customnet,anydoor,dreamedit,lu2023dreamcom,tao2024motioncom,zhang2024zerocomp,winter2024objectdrop,zhang2024inserting,yang2025unicom,liang2025hocomp} require training or finetuning. They can be further divided into two subgroups, according to whether object-specific finetuning is indispensable.  
The first subgroup of methods \cite{PBE,objectstitch,PhDZhang,zhang2023controlcom,yuan2023customnet,anydoor,vamsi2025insert} train a diffusion model on abundant pairs of foregrounds and backgrounds, so that it can be directly applied to a new pair of foreground and background at test time. 
These models need to take one or more reference images of the inserted foreground object as input. 
In the testing period, if a few images containing the foreground object are available, we can optionally finetune the pretrained model on these images and may achieve better performance. The second subgroup of methods \cite{chen2025freecompose,lu2023dreamcom,dreamedit,ruiz2024magic} do not train the model on large-scale dataset. Most of them associate the target object with one rare token, so object-specific finetuning (training on a few images of the same foreground object) is indispensable for these methods. Because the first subgroup is dominant, we will mainly introduce the first subgroup of methods. 

Among the first subgroup of training-based methods, the pioneering works like PbE \cite{PBE} and ObjectStitch \cite{objectstitch} construct massive training triplets of foregrounds, backgrounds, and ground-truth real images based on large-scale image datasets~\cite{Kuznetsova2020TheOI}, in which the foregrounds are cropped from real images followed by color and geometry perturbation. Then, they adapt conditional diffusion model to this task. In particular, the background image, bounding box mask, and  noisy image are concatenated as input, while the foreground is injected into the network via cross-attention.
\citet{humaninsert} adopted a similar approach, but focused on human generation. Some subsequent methods focus on enhancing the ability of detail preservation. For example, \citet{zhang2023controlcom} proposed global-and-local fusion, in which shallow foreground features are used to enhance the details. \citet{anydoor} extracted high-frequency information for better detail preservation. 
\citet{yu2025omnipaint} designed a cycle formed by adding object and removing object, which is expected to facilitate each other.
Recently, inspired by in-context learning, several recent works \cite{wang2025unicombine,song2025insert,huang2025dreamfuse} explored image composition task under the framework of in-context learning. Specifically, noisy latent, background, foreground are equally fed into diffusion model and interact with each other. Different types of input are added with different levels of noise. In-context learning methods demonstrate clear advantage in preserving the subtle foreground details, but they exhibit noticeable copy-and-paste effect and lag in viewpoint/pose adjustment. \cite{osinsert} is a combination of \cite{objectstitch} and \cite{song2025insert}, that is, using \cite{objectstitch} to generate foreground mask and \cite{song2025insert} to fill in the foreground mask. Therefore, \cite{osinsert} can accomplish
both detail preservation and viewpoint/pose adjustment.  

Different works \cite{zhang2023controlcom,yuan2023customnet} also attempted to control image composition from different perspectives. For example,  \cite{yuan2023customnet} provided the target camera viewpoint of foreground object. \cite{zhang2023controlcom} can selectively adjust the illumination and pose of foreground object to match the background. Some methods \cite{he2024affordance,li2024bifr,Zhao2019UnconstrainedFO} aimed to insert the object into any reasonable place in the background image without the provided bounding box. Some methods \cite{Zhao2019UnconstrainedFO,winter2024objectmate} explored generating plausible shadow and reflection for the inserted foreground without the spatial constraint of bounding box.

\subsection{Datasets and Evaluation Metrics}

Training diffusion model requires massive training triplets of foregrounds, backgrounds, and ground-truth real images.  Previous works \cite{PBE,objectstitch} proposed to crop the foregrounds from real images and perturb the foregrounds (\emph{e.g.}, color transfer, geometric transformation), so that we can have perturbed foreground, masked background, and ground-truth real image. 
The multi-view datasets and video datasets can also be used to simulate more diverse and realistic geometry perturbation. In particular, they replace the perturbed foreground with the foreground image from another viewpoint or another video frame. 

Another way to construct image composition dataset is capturing two images with or without the foreground object for the same scene \cite{winter2024objectmate,kim2025orida}. Specifically, they first capture an image for the background scene. Then, they place a foreground object and capture an image with the same camera viewpoint again. However, the scale of such datasets is limited by the high cost of image collection. 

In real-world application scenarios, there exist no ground-truth images for a pair of foreground and background, so we cannot calculate the distance between generated image and ground-truth image. Therefore, previous works  \cite{PBE,objectstitch,PhDZhang} used FID~\cite{FIDHeusel2017} to measure the distribution discrepancy between generated images and real images. Quality metrics \cite{QSscore,wang2004image}  are used to evaluate the authenticity of each image. CLIP score \cite{CLIPscore} or DINO score~\cite{caron2021emerging} are used to measure the similarity between generated foreground and reference foreground. User study should also be conducted to evaluate different aspects for the generated images.

\subsection{Experiments}

We evaluate different methods ObjectStitch~\cite{objectstitch}, InsertAnything~\cite{song2025insert}, OSInsert~\cite{osinsert}, Banana pro~\cite{team2024gemini}, and Seedream 5.0 on MureCom~\cite{lu2023dreamcom} dataset. The visualization results are shown in Fig.~\ref{fig:generative_composition_results}. 

From Fig.~\ref{fig:generative_composition_results}, we can see that \cite{objectstitch} can generate foreground object with reasonable viewpoint/pose but fails to preserve the appearance details. \cite{song2025insert} can preserve the appearance details, but weak in adjusting the foreground viewpoint/pose to match the background when the viewpoint/pose of foreground reference image is not compatible with the background. \cite{osinsert} combines the advantages of \cite{objectstitch} and \cite{song2025insert}, accomplishing both detail preservation and viewpoint/pose adjustment. 

Banana pro and Seedream 5.0 are very competitive commercial models with plausible foreground viewpoint/pose and faithful foreground details, but they have two practical flaws. On the one hand, the generated foregrounds are not well aligned with the provided bounding boxes. Specifically, the inserted objects are often slightly offset, scaled improperly, or partially out of the designated bounding box region. On the other hand, the original background’s color tone and luminance present slight yet discernible alterations in the generated composite images, which disrupts the visual consistency and integrity of the background scene.

%% file: sections/foreground_object_search.tex
\section{Foreground Object Search}\label{sec:foreground_object_search}

\begin{figure}[t]
    \begin{center}
    \includegraphics[width=0.99\linewidth]{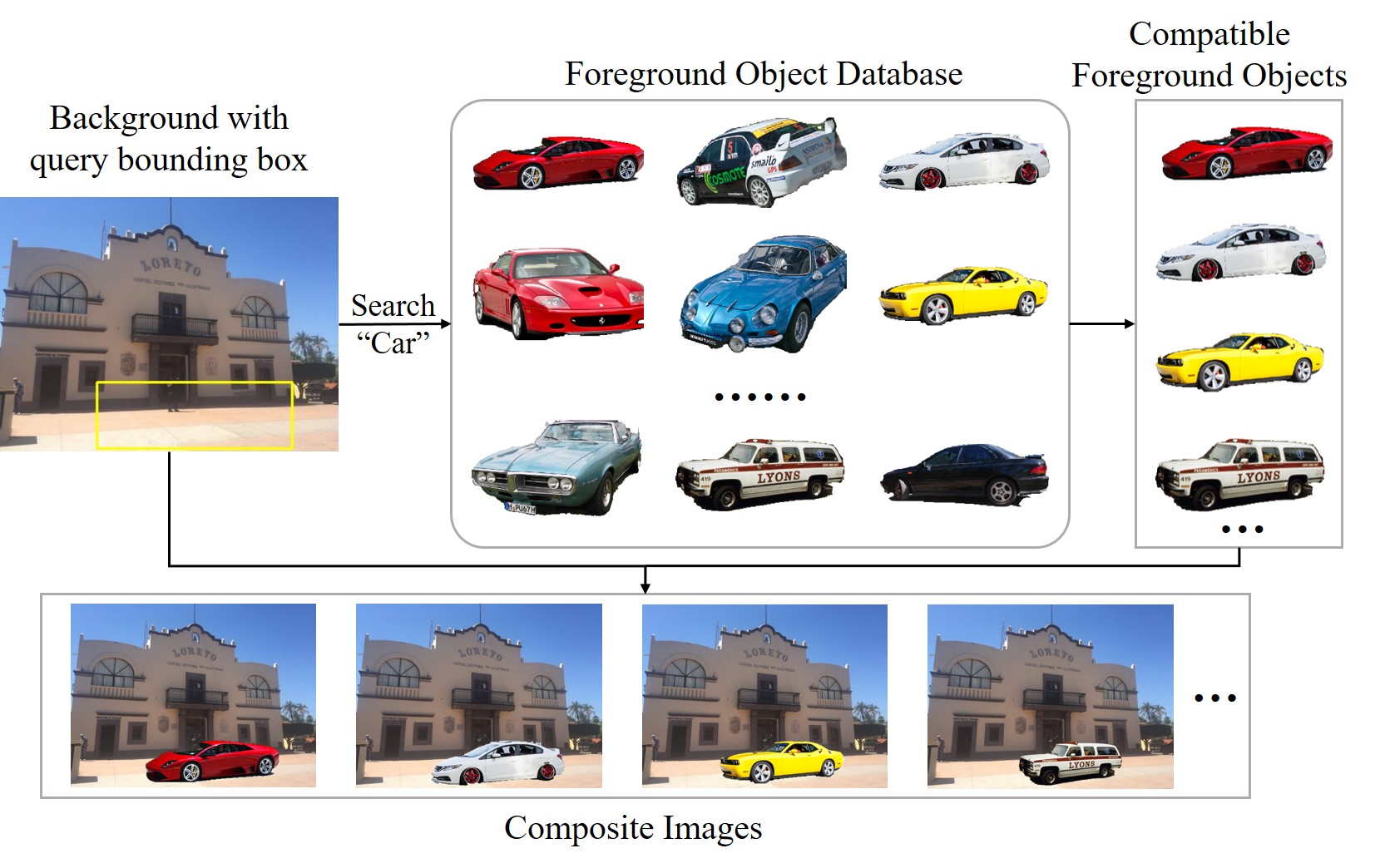}
    \end{center}
    \caption{Illustration of foreground object search. Given a background image with query bounding box (yellow), foreground object search aims to find compatible foreground objects of a specified category from a library, which is combined with the background to produce a realistic composite image.}
    \label{fig:fos_illustration}
\end{figure}

\begin{figure*}[t]
    \begin{center}
    \includegraphics[width=0.99\linewidth]{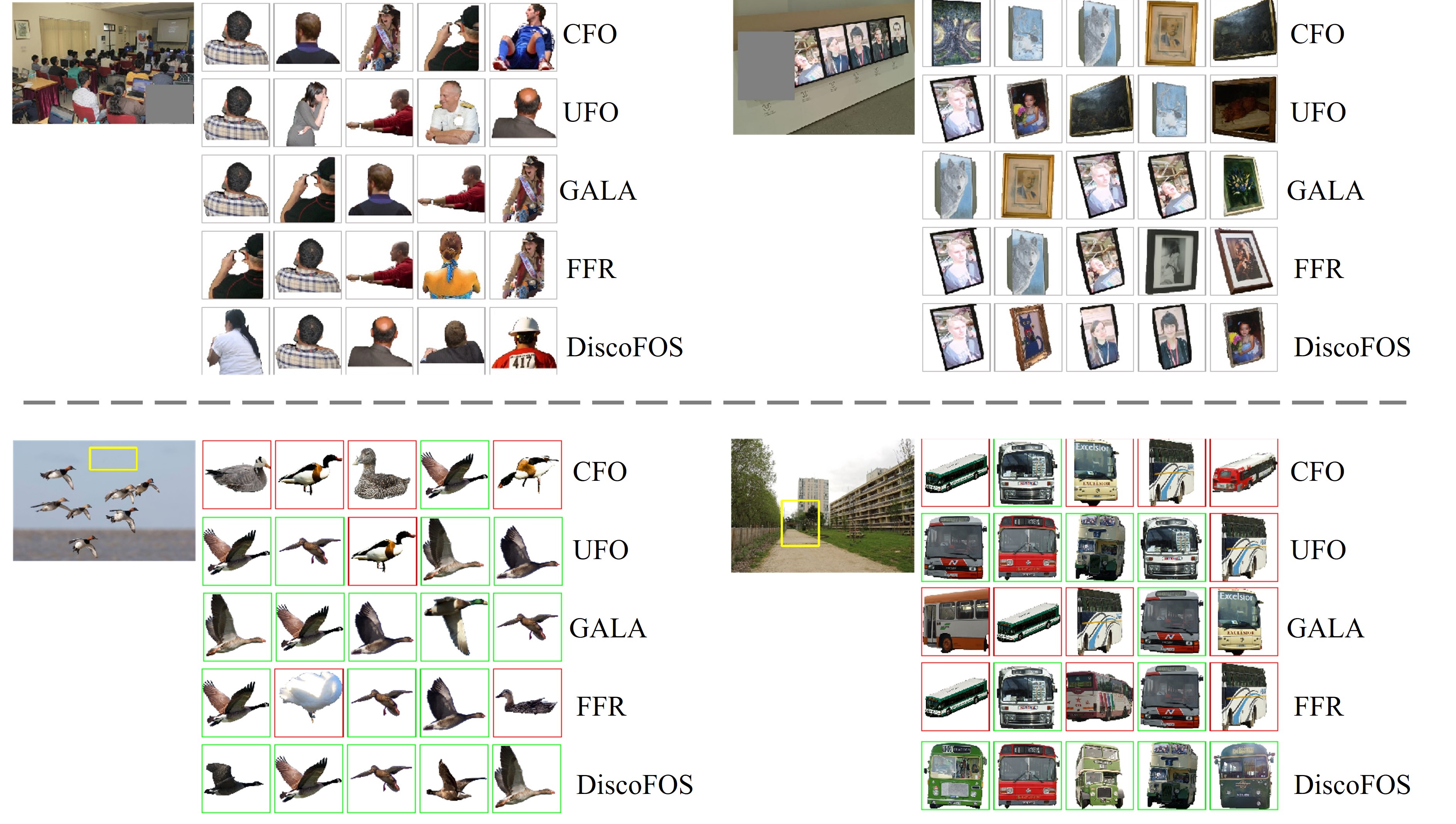}
    \end{center}
    \caption{The visualization results of different foreground object search methods CFO~\cite{Zhao2018CompositingAwareIS}, UFO~\cite{Zhao2019UnconstrainedFO}, GALA~\cite{Zhu2022GALATG}, FFR~\cite{Wu2021FinegrainedFR},  DiscoFOS~\cite{DiscoFOS} on S-FOSD~\cite{DiscoFOS} (top) and R-FOSD~\cite{DiscoFOS} (bottom) datasets. On R-FOSD test set, green (\emph{resp.}, red) box is used to indicate the foreground with compatible (\emph{resp.}, incompatible) label.}
    \label{fig:FOS_results}
\end{figure*}

The goal of foreground object search (FOS) is to retrieve suitable foreground objects from a foreground library, which are compatible with the background in terms of illumination, geometry, and semantics. The FOS task is illustrated in Fig.~\ref{fig:fos_illustration}. 
Finding compatible foreground objects can significantly reduce the effort required to create realistic composite images, complementing other image composition techniques. FOS task can be divided into constrained or unconstrained depending on whether the foreground category is specified. 

\subsection{Traditional Methods}

Early works \cite{Lalonde2007PhotoCA,Chen2009Sketch2PhotoII} attempted to match each foreground with the background using hand-crafted features, but their performance is limited by the expressiveness of hand-crafted features. Specifically, \citet{Lalonde2007PhotoCA} estimated the object information (\emph{e.g.}, size, orientation, lighting condition) and designed matching criteria to rank all the objects in the library. 
\citet{Chen2009Sketch2PhotoII} exploited the contour consistency and content consistency between foreground and background based on hand-crafted features. 

\subsection{Deep Learning Methods}

Recent work used deep learning features for foreground retrieval. For example, \citet{Tan2018WhereAW} utilized deep features to capture local context particularly for person compositing. 
\citet{zhu2015learning} trained a composite image discriminator to predict the realism of composites by compositing each foreground with the background. This method is effective in using the realism of composite images to measure the foreground-background compatibility, but computing the realism of all composite images is very expensive. 
More recent methods~\cite{Zhao2018CompositingAwareIS,Zhao2019UnconstrainedFO,Zhu2022GALATG,Wu2021FinegrainedFR,Li2020InterpretableFO,DiscoFOS} typically trained two encoders to extract foreground feature and background feature. Then, the foreground-background compatibility is measured by calculating the distance between foreground feature and background feature. They share a similar framework, despite the difference in data preparation, network structure, and loss design. \citet{DiscoFOS} observed that a  composite image discriminator \cite{zhu2015learning} can perform much better than two encoders, so they developed a teacher-student network which distills composite image feature from the discriminator to the interaction output of foreground feature and background feature. 

As introduced in Section~\ref{sec:intro}, the foreground and background in a composite image have multiple types of inconsistencies. 
The existing FOS works considered different sets of inconsistencies between background and foreground. For example, the methods~\cite{Zhao2018CompositingAwareIS,Zhao2019UnconstrainedFO} considered the semantic consistency. The methods ~\cite{Li2020InterpretableFO,DiscoFOS} considered the geometric consistency and semantic consistency. Besides the geometric and semantic consistency, some other methods~\cite{Wu2021FinegrainedFR,Zhu2022GALATG} additionally considered style consistency~\cite{Wu2021FinegrainedFR} or illumination consistency~\cite{Zhu2022GALATG}.

\subsection{Datasets and Evaluation Metrics}

Early FOS studies \cite{Zhao2018CompositingAwareIS,Zhao2019UnconstrainedFO,Wu2021FinegrainedFR,Zhu2022GALATG} did not release their datasets. \citet{DiscoFOS} contributed two datasets: S-FOSD and R-FOSD, which contain synthetic composite images and real composite images respectively. 
In S-FOSD dataset, \citet{DiscoFOS} segment one foreground object from a real image and fill its bounding box with image mean values to get the background. For each background image, the foreground object from the same image is deemed as ground-truth. 
In R-FOSD dataset, \citet{DiscoFOS} collect images from Internet as background images and draw a bounding box at the expected foreground location as query bounding box. R-FOSD dataset uses the same foregrounds with the test set of S-FOSD dataset. \citet{DiscoFOS} employ human annotators to label the compatibility of each pair of background and foreground.
In comparison, S-FOSD dataset is low-cost and highly scalable, but does not contain complete background images or ground-truth negative samples. R-FOSD dataset has complete background image with accurately annotated positive and negative foregrounds, but is unscalable due to the high annotation cost.

On the synthetic composite image dataset, Recall@k (R@k) is adopted as an evaluation metric~\cite{Zhao2018CompositingAwareIS,Zhu2022GALATG,DiscoFOS}, which represents the percentage of background queries whose ground-truth foreground appears in top $k$ retrievals.
On the real composite image dataset, mean Average Precision (mAP), mAP@20, and Precision@k (P@k) are adopted as an evaluation metrics~\cite{Zhao2019UnconstrainedFO,Zhu2022GALATG,DiscoFOS}.

\subsection{Experiments}

We evaluate different methods on S-FOSD dataset and R-FOSD dataset~\cite{DiscoFOS}. Specifically, we train on S-FOSD training set, while testing on S-FOSD and R-FOSD test sets. The retrieval results of  CFO~\cite{Zhao2018CompositingAwareIS}, UFO~\cite{Zhao2019UnconstrainedFO}, GALA~\cite{Zhu2022GALATG}, FFR~\cite{Wu2021FinegrainedFR}, and DiscoFOS~\cite{DiscoFOS} are shown in Fig.~\ref{fig:FOS_results}. In each example, we show the background image with query bounding box (gray square or yellow box) on the left side and top five retrieved results of different methods on the right side.  
The results show that DiscoFOS can retrieve more foregrounds that are compatible with the background geometrically and semantically.

%% file: sections/conclusion.tex
\section{Conclusion}\label{sec:conclusion}

In this paper, we have conducted a comprehensive survey on image composition, which involves a variety of techniques to produce a realistic composite image. We have introduced object placement, image blending, image harmonization, shadow generation, generative image composition, and foreground object search. In the future, we will extend this survey to broader composition tasks in related fields, such as video composition and 3D/4D composition.